\documentclass[]{cit_lab_mfr}

\usepackage{amssymb}
\usepackage{amsmath}
\usepackage{mathtools}
\usepackage{enumitem}
\usepackage{graphicx}
\usepackage{booktabs}
\usepackage{multirow}
\usepackage{tabularx}
\usepackage{float}
\usepackage{placeins}
\usepackage{algorithm}
\usepackage{algpseudocode}
\usepackage{capt-of}
\usepackage{pifont}
\usepackage{xurl}
\definecolor{markgreen}{HTML}{008300}
\definecolor{markred}{HTML}{E34948}
\newcommand{\cmark}{\textcolor{markgreen}{\ding{51}}}
\newcommand{\xmark}{\textcolor{markred}{\ding{55}}}

\title{{\fontsize{15.5}{20}\selectfont UniWAM Technical Report: Unified Mobile Manipulation via Mixed-Stream World-Action Modeling and Manipulation Anchor Pose Supervision}}

\author{%
\parbox{\textwidth}{\centering
Wei Xue$^{*}$, Keliang Liu$^{*}$, Mingzhang Cui, Jinhua Xie, Jinjie Wei, Jianan Hou, Jingcheng Lu, \\
Lintao Wang, Kaixiang Qiu, Yizhou Liu, Xinghai Ye, Jinghang Han, Mingcheng Li, Jie Gu, \\
Shunli Wang, Lihua Zhang$^{\S}$, Dingkang Yang$^{\dagger,\S}$
}}

\affiliation{%
\parbox{\textwidth}{\centering\small
Physical Superintelligence Lab, Fysics AI \qquad College of Intelligent Robotics and Advanced Manufacturing, Fudan University
}}

\contribution[*]{Equal contribution; order decided by a coin flip}
\contribution[\dagger]{Project lead}
\contribution[\S]{Corresponding author}

\abstract{
Mobile manipulation requires precise navigation to a manipulation-ready pose followed by reliable object interaction. These two stages differ in action spaces and visual requirements, which complicates unified policy learning. In addition, collecting diverse real-world navigation data with explicit manipulation-ready pose supervision remains costly and difficult to scale. We introduce UniWAM, a unified mixed-stream world-action model with separate action encoders and output heads for navigation and manipulation, sharing a common backbone. This design supports joint representation learning on independently sampled navigation and manipulation data. UniWAM supports independent inference for either stream and batch-parallel inference for both. We further introduce Manipulation Anchor Pose (MAP) supervision for where to stop and how to orient for manipulation. An automated pipeline constructs MAP-Data from large-scale 3D scenes, yielding over 1.5 million episodes and 7,500 hours. MAP-Data provides per-frame target-object bounding boxes and image-plane MAP coordinates as auxiliary navigation supervision. Together with projected end-effector trajectories for manipulation, these prediction targets provide stream-specific image-plane supervision for action learning from egocentric observations. With large-scale MAP-Data, UniWAM outperforms the strongest external baselines on our MAP-Bench by 30.1\% in position error and 44.0\% in heading error. Across 24 real-robot tasks, UniWAM achieves leading results in MAP navigation and mobile manipulation, with competitive manipulation performance. We have released code, data, and benchmark.
}
\date{September 3, 2026}
\checkdata[Corresponding]{\begin{minipage}[t]{0.8\linewidth}\raggedright\hyphenpenalty=10000 wxue24@m.fudan.edu.cn, klliu25@m.fudan.edu.cn, dkyang20@fudan.edu.cn, lihuazhang@fudan.edu.cn\end{minipage}}
\checkdata[Project Page]{\url{https://fysics-ai.github.io/uniwam.github.io/}}
\checkdata[Github]{\url{https://github.com/Fysics-AI/UniWAM}}
\checkdata[Hugging Face]{\begin{minipage}[t]{0.8\linewidth}\raggedright \url{https://huggingface.co/Fysics-AI/UniWAM}\\ \url{https://huggingface.co/datasets/Fysics-AI/MAP-Data}\\ \url{https://huggingface.co/datasets/Fysics-AI/MAP-Bench}\end{minipage}}

\begin{document}
\maketitle

\section{Introduction}
\label{sec:intro}

\begin{figure}[t]
\centering
\includegraphics[width=\textwidth]{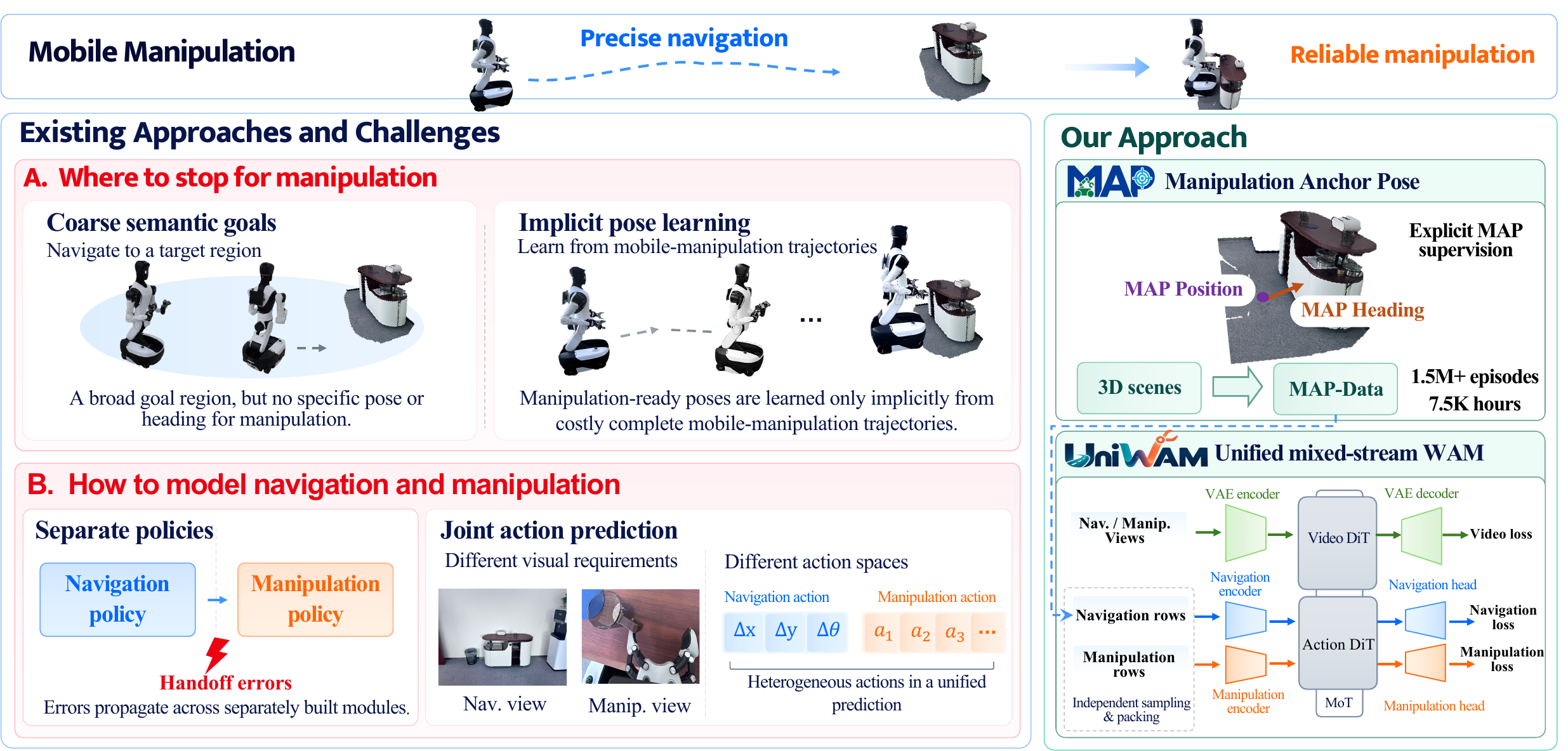}
\caption{Challenges of mobile manipulation and our approach. Existing unified policies either share one action representation across navigation and manipulation or separate the two only inside the action module, and they learn where to stop only implicitly from costly complete trajectories. UniWAM models navigation and manipulation as separate streams over a shared backbone, and MAP-Data supervises the manipulation-ready pose at scale with automatically generated trajectories.}
\label{fig:motivation-overview}
\end{figure}

Mobile manipulation requires a robot to move through an environment and interact precisely with objects after arrival. Compared with navigation or fixed-base manipulation alone, mobile manipulation supports a much broader range of tasks, such as fetching an item from another room or clearing a table in a different part of a house~\citep{yenamandra2023homerobot,fu2024mobilealoha}. However, navigation and manipulation differ in both what they need to observe and how they act. Navigation needs a forward view of the scene ahead, including distant targets and free space, whereas manipulation needs detailed views of the local workspace, often from downward-facing and wrist cameras~\citep{yang2026panovla,zhu2026geohat,chen2026mopa}. Their action signals also differ in frequency. Base motion varies slowly and smoothly over long distances, whereas arm and gripper motion changes rapidly near contact~\citep{zhu2026geohat}. A single policy for both behaviors must reconcile two visual requirements and two action spaces.

Research on mobile manipulation mainly follows two routes, namely modular systems and unified models. Modular systems train navigation and manipulation policies separately and connect them at execution time~\citep{wu2025momanipvla}. This handoff adds an interface between the two policies and can propagate errors from one module to the next~\citep{zhu2026geohat,wu2025momanipvla}. Unified models instead learn the complete mobile-manipulation process end to end within a single policy. Recent unified approaches build on vision-language-action (VLA) models~\citep{physicalintelligence2026pi07,bjorck2025gr00t} and world-action models (WAMs)~\citep{yuan2026fastwam,fan2026mobilewam} to predict navigation and manipulation actions within a shared model and a shared training pipeline.

Unified mobile manipulation still faces two linked challenges, illustrated in Fig.~\ref{fig:motivation-overview}. The first is architectural. A shared action token for navigation and manipulation~\citep{black2025pi05,bjorck2025gr00t} ignores their heterogeneous visual observations, action spaces, and control rates. Separate action experts~\citep{chen2026abot} decouple the action spaces, yet both designs still jointly supervise both action targets within the same training sample. Because mobile manipulation is largely sequential, the arm is mostly idle during navigation and the base is mostly idle during manipulation. Joint supervision thus spends much of the training on stationary commands, ties the sampling of the two behaviors to complete trajectories, and prevents direct use of navigation-only or manipulation-only data, which are far easier to collect.

The second challenge concerns data and follows from the first. Existing unified models learn the manipulation-ready terminal pose only implicitly from complete trajectories, so they require large-scale mobile-manipulation data, which are costly to collect~\citep{xu2026hommi,zhu2025emma}. Without such data, a policy may reach the target but stop at an unsuitable position or heading, and the subsequent manipulation then fails~\citep{yang2025mobipi}. Semantic navigation data scale better, but they specify only target proximity, not the terminal pose required for manipulation~\citep{internvla2025}.

To address the first challenge, we propose \textbf{UniWAM}, a mixed-stream world-action model with separate navigation and manipulation streams. Each stream has its own visual inputs, action encoder, and output head over shared video and action DiTs. The two streams are sampled independently and packed as separate batch rows without cross-row attention, so each sample supervises only its own stream. Navigation-only, manipulation-only, and mobile-manipulation data can all be used directly, and either stream runs alone or both run in one batch at deployment.

To address the second challenge, we introduce the \textbf{Manipulation Anchor Pose (MAP)}, the base position and heading at which navigation should end for manipulation. Unlike floor affordances, candidate base placements, or pose preferences learned from rollouts~\citep{zhang2025momakitchen,zhong2026floaff,chen2026gbpp,chai2025n2m}, MAP is an explicit target that can be generated automatically and supervises the entire approach trajectory. We build \textbf{MAP-Data} from large-scale indoor 3D scenes with an automated pipeline, yielding over 1.5 million navigation episodes and 7,500 hours annotated with per-frame target bounding boxes and image-plane MAP coordinates. We hold out a subset of its targets as \textbf{MAP-Bench} for evaluating MAP navigation. Because the streams are sampled independently, UniWAM trains its navigation stream directly on MAP-Data, which supervises where to stop for mobile manipulation explicitly and at scale. Together with projected end-effector trajectories for manipulation, the bounding boxes and MAP coordinates serve as image-plane auxiliary targets for the two streams~\citep{tu2026sgvla,yang2026dreamtrajectory}.

Our contributions are:
\begin{itemize}[leftmargin=*,itemsep=2pt,topsep=4pt]
    \item We propose \textbf{UniWAM}, a mixed-stream world-action model that learns navigation and manipulation as two independently sampled streams over a shared backbone.
    \item We define the \textbf{Manipulation Anchor Pose (MAP)} and generate \textbf{MAP-Data}, over 1.5M episodes labeled with MAP and target boxes. To our knowledge, it is the largest dataset for manipulation-oriented navigation. We hold out \textbf{MAP-Bench}, a leakage-free benchmark for MAP navigation.
    \item We evaluate UniWAM on MAP-Bench and 24 real-robot tasks. Through mixed training on MAP-Data and diverse manipulation data, UniWAM outperforms pretrained baselines, reducing position and heading errors by 30.1\% and 44.0\% and raising mobile-manipulation success by 17.5 points.
\end{itemize}

\section{Related Work}
\label{sec:related}

\textbf{Unified mobile-manipulation models.} Most unified policies predict base, arm, and gripper commands through a shared action representation. RT-1 adds base motion to arm control and selects the controlled component with a discrete mode variable~\citep{brohan2023rt1}. Mobile ALOHA concatenates bimanual joint positions with base linear and angular velocities into one action vector~\citep{fu2024mobilealoha}. VLA models such as $\pi_0$~\citep{black2024pi0}, $\pi_{0.5}$~\citep{black2025pi05}, GR00T~\citep{bjorck2025gr00t}, and Xiaomi-Robotics-1~\citep{xiaomi2026robotics1} follow the same interface. They predict all degrees of freedom in one action chunk and use navigation-only or manipulation-only data only by zero-padding or masking the absent dimensions. This design learns coordinated behavior, but one action token must cover navigation and manipulation despite their different visual inputs, action spaces, and control rates, and each sample from a complete trajectory supervises both behaviors, including the idle one. World-action models (WAMs) build on video generation models to jointly predict future visual states and actions~\citep{ye2026dreamzero,kim2026cosmospolicy,pai2025mimicvideo,bi2025motus,li2026lingbotva,yuan2026fastwam}. Most WAMs target fixed-base manipulation. ABot-M0.5~\citep{chen2026abot} and MobileWAM~\citep{fan2026mobilewam} extend them to mobile manipulation. ABot-M0.5 assigns base and manipulation actions to separate sub-towers within a dual-level mixture-of-transformers. MobileWAM attaches an action expert to a pretrained video diffusion transformer, routes its feed-forward layers across shared, locomotion, and manipulation experts, and supervises future latents with Chain-of-Foresight. Both methods separate navigation and manipulation only within the action module. The two behaviors still share visual inputs, and each sample still supervises both. Cross-embodiment policies such as CrossFormer~\citep{doshi2024crossformer} and HPT~\citep{wang2024hpt} use embodiment-specific tokenizers or stems and action heads over a shared trunk, so each sample supervises the action space of one robot. They separate different robots, however, not the navigation and manipulation phases of one mobile manipulator, and they have neither a video world model nor an explicit terminal pose. UniWAM applies this separation within one mobile manipulator. Each stream has its own visual inputs, action encoder, and output head and is sampled independently, while both streams share the video and action DiTs. It further combines this design with a video world model and explicit MAP supervision. Table~\ref{tab:new-unified} summarizes these differences across views, sampling, world modeling, and terminal-pose supervision.

\begin{table}[t]
\caption{Comparison with recent unified mobile-manipulation models and cross-embodiment policies. The first three are VLA policies, the next two are world-action models, and the last two are cross-embodiment policies, which separate different robots rather than the navigation and manipulation phases of one mobile manipulator. Separate views: navigation and manipulation receive different camera inputs. One behavior per sample: a training sample can supervise one behavior alone without padding or masking. Video world model: the model predicts future video. Explicit terminal pose: the terminal base pose for manipulation is supervised directly.}
\label{tab:new-unified}
\centering
\footnotesize
\setlength{\tabcolsep}{4pt}
\resizebox{\linewidth}{!}{%
\begin{tabular}{@{}llcccc@{}}
\toprule
Model & Action generator & \shortstack{Separate\\views} & \shortstack{One behavior\\per sample} & \shortstack{Video world\\model} & \shortstack{Explicit\\terminal pose} \\
\midrule
$\pi_{0.5}$~\citep{black2025pi05} & single action expert for all degrees of freedom & \xmark & \xmark & \xmark & \xmark \\
GR00T N1.7~\citep{bjorck2025gr00t} & single action head for all degrees of freedom & \xmark & \xmark & \xmark & \xmark \\
Xiaomi-Robotics-1~\citep{xiaomi2026robotics1} & single action head for all degrees of freedom & \xmark & \xmark & \xmark & \xmark \\
ABot-M0.5~\citep{chen2026abot} & action stream with mobility and manipulation sub-towers & \xmark & \xmark & \cmark & \xmark \\
MobileWAM~\citep{fan2026mobilewam} & action expert with routed locomotion and manipulation experts & \xmark & \xmark & \cmark & \xmark \\
CrossFormer~\citep{doshi2024crossformer} & shared transformer with embodiment-specific action heads & \cmark & \cmark & \xmark & \xmark \\
HPT~\citep{wang2024hpt} & shared trunk with embodiment-specific stems and heads & \cmark & \cmark & \xmark & \xmark \\
\midrule
UniWAM & shared action DiT with separate stream interfaces & \cmark & \cmark & \cmark & \cmark \\
\bottomrule
\end{tabular}}
\end{table}

\textbf{Manipulation-oriented navigation.} Unlike semantic navigation, which only requires reaching the vicinity of a target~\citep{zhang2024uni,internvla2025}, manipulation-oriented navigation must find a base position and heading that support the next interaction. Existing methods fall into three groups. The first searches for feasible base placements from arm reachability~\citep{vahrenkamp2013robot}. The second learns manipulation-ready locations from labeled or executed data, including floor affordance maps~\citep{zhang2025momakitchen,zhong2026floaff}, grasp-aware base placement~\citep{chen2026gbpp}, and pose preferences learned from manipulation rollouts~\citep{chai2025n2m}. The third infers the base pose at test time by optimizing compatibility with a manipulation policy~\citep{yang2025mobipi} or by reasoning with vision-language models~\citep{wu2025moto,zhang2026unilmnav}. These methods output candidate poses, affordance regions, or placement scores for a specific robot, policy, or scene. None of them supervises the approach trajectory that leads to the chosen pose. As a result, navigation data that explicitly supervise the manipulation-ready terminal pose along entire approach trajectories remain scarce, although such supervision is essential for learning where to stop. We define this terminal pose as the MAP and generate MAP-Data from large-scale indoor 3D scenes. MAP-Data supervises the entire approach with an explicit terminal pose and dense image-plane labels. The largest prior dataset, MoMa-Kitchen with 127K episodes, is about one-twelfth the size of MAP-Data with its 1.5M episodes. Table~\ref{tab:new-placement} compares them in output, label source, and scale.

\begin{table}[t]
\caption{Comparison of MAP-Data with methods for manipulation-ready base placement. Supervises approach: the labels cover the entire approach trajectory rather than only a terminal pose or a score for a given scene. Scale is the amount of labeled data reported in each paper, and a dash denotes methods that do not learn from a dataset, such as kinematic search or test-time optimization.}
\label{tab:new-placement}
\centering
\footnotesize
\setlength{\tabcolsep}{4pt}
\resizebox{\linewidth}{!}{%
\begin{tabular}{@{}lllcc@{}}
\toprule
Method & Output & Label source & \shortstack{Supervises\\approach} & Scale \\
\midrule
Reachability inversion~\citep{vahrenkamp2013robot} & feasible base poses & arm kinematics & \xmark & -- \\
MoMa-Kitchen~\citep{zhang2025momakitchen} & floor affordance map & simulation & \xmark & 127K episodes \\
FloAff-Kitchen~\citep{zhong2026floaff} & floor affordance map & simulation & \xmark & 24.8K observations \\
GBPP~\citep{chen2026gbpp} & scores of candidate base poses & rule labels and simulation trials & \xmark & 180K rule and 12K simulation labels \\
N2M~\citep{chai2025n2m} & distribution over base poses & policy rollouts & \xmark & 12 to 15 rollouts per task \\
Mobi-$\pi$~\citep{yang2025mobipi} & base pose for a given policy & test-time optimization & \xmark & -- \\
\midrule
MAP-Data & terminal pose, per-frame box and MAP labels & scene geometry and a VLM & \cmark & 1.5M episodes \\
\bottomrule
\end{tabular}}
\end{table}

\section{Method}
\label{method}

This section first defines the MAP and describes how MAP-Data and MAP-Bench are built (Sec.~\ref{sec:map}). It then presents UniWAM, including its stream interfaces, mixed-stream batch packing, independent stream sampling, image-plane auxiliary supervision, training objective, and implementation (Sec.~\ref{sec:uniwam}).

\begin{figure}[t]
\centering
\includegraphics[width=\textwidth]{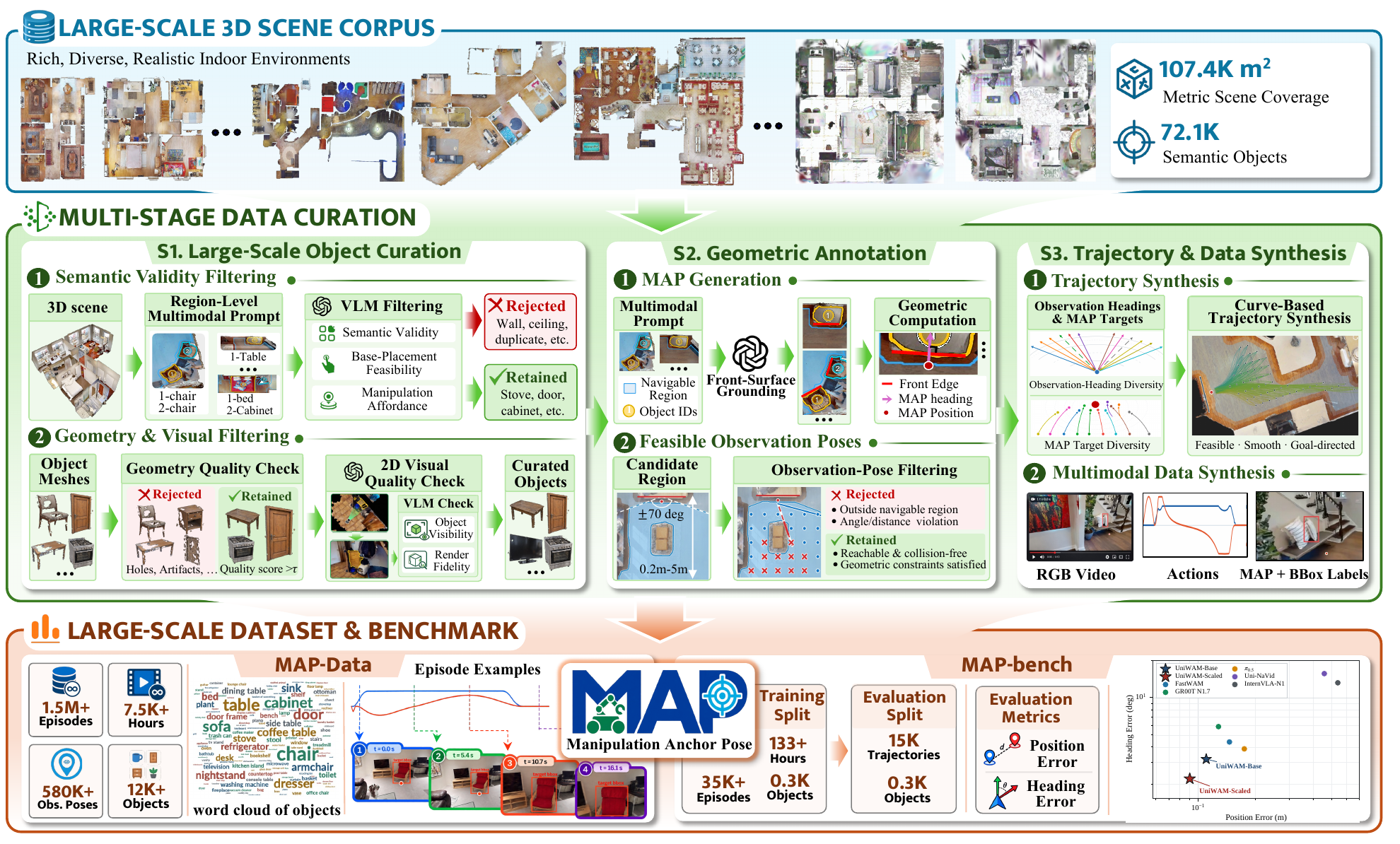}
\caption{Overview of MAP-Data and MAP-Bench. From large-scale indoor 3D scenes, S1 curates valid target objects, S2 annotates each target with a MAP and feasible observation poses, and S3 synthesizes approach trajectories with RGB video, actions, and MAP and box labels. MAP-Bench evaluates position and heading errors from held-out initial poses.}
\label{fig:map}
\end{figure}

\subsection{MAP: Manipulation Anchor Pose}
\label{sec:map}

Existing representations of manipulation-ready base placement are specific to a robot, policy, or scene and do not supervise the approach trajectory~\citep{zhang2025momakitchen,chai2025n2m}. We instead make the terminal base pose an explicit target and generate approach trajectories toward it at scale, which yields dense supervision for every step of the approach.

\textbf{MAP definition.} For a manipulation target $o_i$, its \emph{Manipulation Anchor Pose} (MAP) is $\mathbf m_i=(\mathbf p_i^{\mathrm{MAP}},\theta_i^{\mathrm{MAP}})$, where $\mathbf p_i^{\mathrm{MAP}}\in\mathbb R^2$ and $\theta_i^{\mathrm{MAP}}\in\mathbb S^1$ denote the base position and heading. The MAP explicitly specifies where the robot should stop and how it should orient for manipulation. Because it is defined by scene geometry rather than by a particular manipulation policy, it can be computed automatically for any target in a 3D scene.

\textbf{Scene sources and target curation.} MAP-Data is built from HM3D and HM3D-Semantics~\citep{ramakrishnan2021hm3d,yadav2022habitat}, InteriorGS~\citep{miao2026towards}, and a large in-house scene corpus. As illustrated in Fig.~\ref{fig:map}, the pipeline has three stages. Stage S1 converts these heterogeneous sources into a common representation of target instances, navigable regions, and renderable observations. Mesh-based scenes undergo geometry and navigability checks, whereas Gaussian-splatting scenes are validated through occupancy maps and rendered observations. For sources without object-instance annotations, semantic instance segmentation supplies candidate target masks. Candidate targets then pass four filters. Semantic filtering verifies that the instance matches the task condition and is uniquely identifiable. Geometry filtering rejects incomplete target surfaces and corrupted scene regions. Base-placement filtering requires collision-free navigable space near the manipulation-facing surface. Visibility and rendering filtering removes targets that cannot be observed reliably from feasible camera poses or that produce invalid rendered views, and the remaining targets proceed to MAP generation.

\textbf{MAP and observation-pose generation.} Stage S2 constructs the MAP of each target and a set of feasible observation poses. A vision-language model identifies the manipulation-facing front edge of the target. Let $\mathbf p_f$ be the ground-plane midpoint of this edge and $\mathbf n_f$ its unit normal pointing toward navigable space. Denoting the clearance-aware navigable region by $\mathcal F_{\mathrm{nav}}$, the MAP position and heading are
\begin{equation}
\mathbf p^{\mathrm{MAP}}=\operatorname{Ray}(\mathbf p_f,\mathbf n_f)\cap \partial\mathcal F_{\mathrm{nav}},
\qquad
\theta^{\mathrm{MAP}}=\operatorname{atan2}\!\left((\mathbf p_f-\mathbf p^{\mathrm{MAP}})_y,(\mathbf p_f-\mathbf p^{\mathrm{MAP}})_x\right).
\label{eq:map-ray-construction}
\end{equation}
The intersection of the ray with the boundary of the navigable region determines where the base stops, and the heading makes the robot face the target surface. Observation poses are sampled independently across target distance, bearing, and camera heading. They are retained only when the base lies in navigable free space, satisfies collision and geometric constraints, and provides a valid view of the target. Sampling observation poses separately from the MAP diversifies the initial states while keeping the terminal target fixed for all trajectories of a target.

\textbf{Trajectory synthesis.} Stage S3 connects each retained observation pose to its MAP by a single-segment or multi-segment cubic B\'ezier trajectory. A segment with control points $\{\mathbf c_j\}_{j=0}^{3}$ is
\begin{equation}
\mathbf q(s)=(1-s)^3\mathbf c_0+3(1-s)^2s\,\mathbf c_1+3(1-s)s^2\mathbf c_2+s^3\mathbf c_3,\qquad s\in[0,1].
\label{eq:map-bezier}
\end{equation}
The control points preserve the initial motion direction and terminate at the MAP with the prescribed heading. If a direct curve is obstructed, A*~\citep{hart1968formal} provides a collision-free corridor that guides piecewise fitting. A trajectory is retained only if it remains within the clearance-aware navigation mask, approaches the target from the annotated side, keeps the manipulation-facing boundary within the horizontal field of view, and satisfies curvature and path-length constraints. Trajectories with excessive backtracking or an inaccurate terminal position or heading are rejected.

\textbf{Synchronized geometric labels.} At each trajectory step, the renderer records the egocentric RGB observation and the planar navigation action. The MAP position is projected into the current camera as
\begin{equation}
\mathbf u_t^{\mathrm{MAP}}=\Pi\!\left(T_{c_t\leftarrow w}\begin{bmatrix}x_{\mathrm{MAP}} & y_{\mathrm{MAP}} & z_{\mathrm{MAP}} & 1\end{bmatrix}^{\!\top}\right),
\label{eq:map-image-projection}
\end{equation}
where $T_{c_t\leftarrow w}$ maps world coordinates to the camera frame and $\Pi$ applies the camera intrinsics and perspective division. Target-instance masks are converted to image-plane bounding boxes. RGB frames, navigation actions, projected MAP coordinates, target boxes, scene and target identifiers, and quality flags are stored with the same timestamps for frame-level alignment. Each trajectory also carries five target-specific instructions. MAP-Data contains over \textbf{1.5M} episodes and \textbf{7.5K} hours of navigation trajectories with explicit MAP and image-plane geometric supervision.

\textbf{MAP-Bench.} MAP-Bench contains 300 targets in 109 scenes. Its training split contains 35,454 trajectories from 6,732 observation poses, with 7,184,098 RGB frames at 15\,Hz, and its evaluation split contains 15K trajectories. Each target has five equivalent English instructions, and one of them is sampled for each episode with a fixed random seed. Each episode specifies a reference MAP and an initial pose. Episodes are grouped by their initial distance to the reference MAP into a short range below 1.5\,m, a mid range from 1.5 to 3.5\,m, and a long range of at least 3.5\,m. To prevent leakage between the splits, we cluster feasible initial poses in $SE(2)$ with the normalized distance
\begin{equation}
d(\mathbf x_i,\mathbf x_j)=\sqrt{\left(\frac{\|\mathbf p_i-\mathbf p_j\|_2}{r_p}\right)^2+\left(\frac{\operatorname{wrap}(\theta_i-\theta_j)}{r_\theta}\right)^2},
\label{eq:map-se2-distance}
\end{equation}
where $r_p$ and $r_\theta$ set the spatial and angular clustering scales. Each cluster is assigned entirely to one split. The two splits therefore share targets but never share an initial pose or a near-duplicate initial view, so the evaluation measures generalization to unseen starting configurations rather than memorized trajectories. All MAP-Bench targets are further excluded from the rest of MAP-Data, so the additional data used by UniWAM-Scaled contain no benchmark target, although MAP-Data may contain other targets in the benchmark scenes. Sec.~\ref{app:new-zeroshot} evaluates unseen scenes and targets zero-shot on real robots, which complements the held-out initial poses of MAP-Bench.

\textbf{Metrics and evaluation.} We evaluate MAP navigation with the MAP Position Error and the MAP Heading Error. For $N$ evaluation trajectories, they are
\begin{equation}
E_{\mathrm{pos}}=\frac{1}{N}\sum_{r=1}^{N}\left\|\widehat{\mathbf p}_r-\mathbf p_r^{\mathrm{MAP}}\right\|_2,
\qquad
E_{\mathrm{head}}=\frac{1}{N}\sum_{r=1}^{N}\left|\operatorname{wrap}\!\left(\widehat{\theta}_r-\theta_r^{\mathrm{MAP}}\right)\right|,
\label{eq:map-metrics}
\end{equation}
where $\widehat{\mathbf p}_r$ and $\widehat{\theta}_r$ are the reached position and heading, and $\mathbf p_r^{\mathrm{MAP}}$ and $\theta_r^{\mathrm{MAP}}$ are those of the reference MAP. Eq.~\ref{eq:map-metrics} is applied per distance range, and Overall averages the three ranges. All models are evaluated in closed loop from the same held-out initial poses with the same instructions. A predicted action chunk is regarded as near zero if at least 80\% of its predicted relative poses, including the last one, satisfy $\|(x_t,y_t)\|_2\leq0.05$\,m and $|\theta_t|\leq0.05$\,rad. For models that predict velocities, the test is applied to the base displacement integrated over the chunk. An episode stops after three consecutive near-zero chunks, and the third chunk is not executed. Uni-NaVid stops when it predicts its stop token. An episode also ends after twice the steps of its reference trajectory, which bounds the evaluation time of models that never stop.

\begin{figure}[t]
\centering
\includegraphics[width=\textwidth]{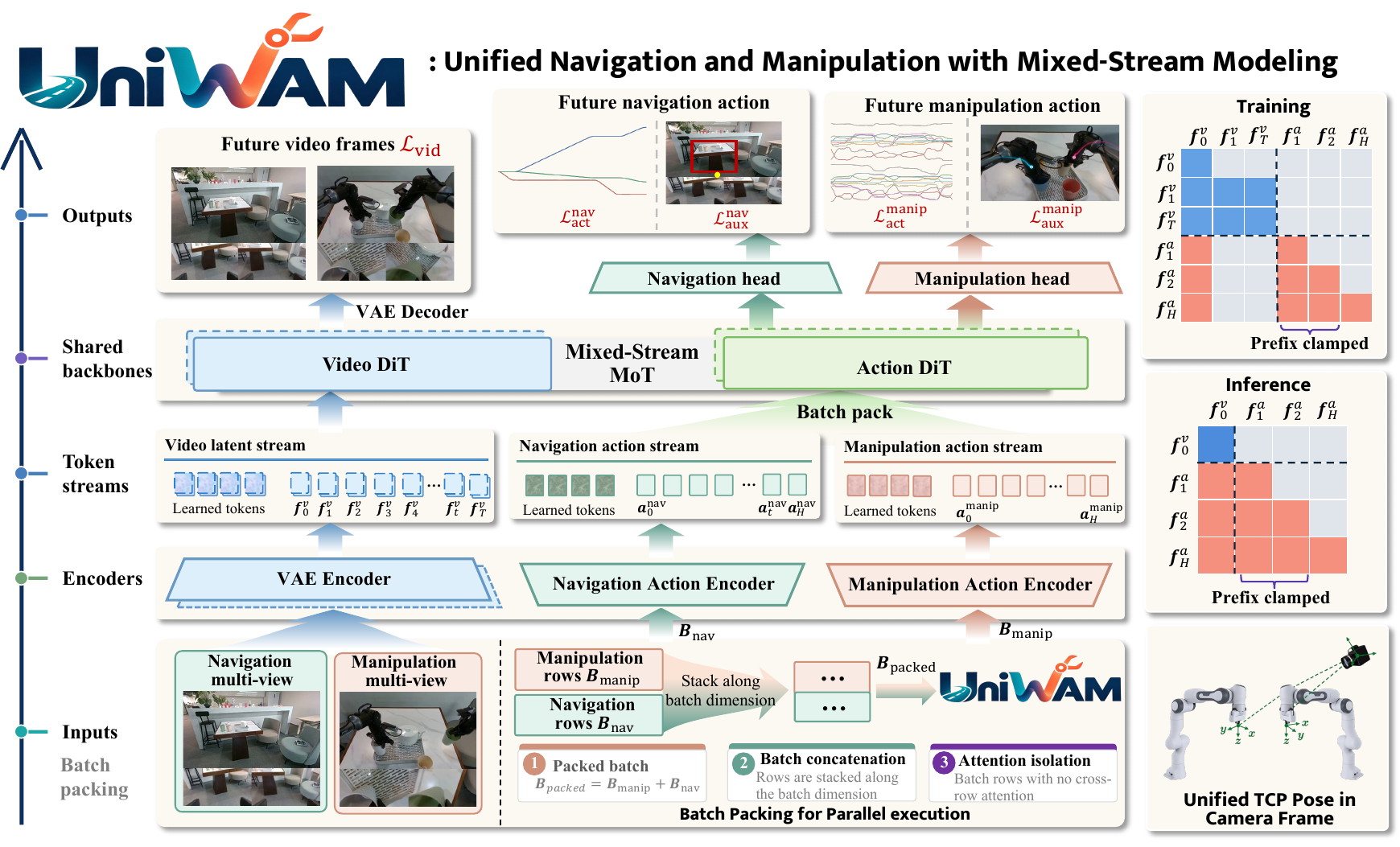}
\caption{Overview of UniWAM. Navigation and manipulation samples are drawn independently, encoded by stream-specific action encoders, and packed as separate batch rows through the shared video and action DiTs. Stream-specific heads predict actions and image-plane auxiliary targets, while the video branch predicts future frames. Right: attention masks with clean action prefixes for training and inference, and the camera-frame end-effector pose shared across embodiments.}
\label{fig:uniwam}
\end{figure}

\subsection{UniWAM: Unified Mixed-Stream World-Action Model}
\label{sec:uniwam}

UniWAM separates navigation and manipulation at the level of control streams rather than within the action module, as shown in Fig.~\ref{fig:uniwam}. Each stream has its own visual inputs, action encoder, output head, and learnable tokens. Both streams share a video DiT, initialized from Wan2.2-TI2V-5B~\citep{wan2025wan}, and an action DiT, coupled through mixture-of-transformers attention~\citep{liang2024mot}. This design addresses the two problems identified in Sec.~\ref{sec:intro}. Stream-specific interfaces accommodate the heterogeneous observations and action spaces, and independent sampling lets each training sample supervise only one stream. The shared DiTs keep a single model and a shared representation for both streams.

\subsubsection{Stream Interfaces and Action Representations}
\label{app:action-representations}

Let $s\in\{\mathrm{nav},\mathrm{manip}\}$. At time $t$, stream $s$ receives camera views $o_t^s$, the instruction $\ell$, and the proprioceptive state $q_t$, and predicts an action chunk
\begin{equation}
A_t^s=(a_{t,1}^s,\ldots,a_{t,H}^s)=f_\phi^{s}\big(o_t^s,\ell,q_t\big),
\label{eq:uniwam-stream}
\end{equation}
where $H$ is the action horizon, a hyperparameter, and $f_\phi^{s}$ shares all DiT parameters $\phi$ across streams. A stream-specific linear encoder maps each action to the shared hidden space, and a stream-specific head decodes it. Because the DiTs are shared, learnable tokens for the video and actions of each stream mark the view and action space of each sequence.

\textbf{Navigation actions.} Many unified policies represent base motion by linear and angular velocities $(v,\omega)$, for example Mobile ALOHA~\citep{fu2024mobilealoha}. UniWAM instead predicts planar pose increments $(\Delta x,\Delta y,\Delta\theta)$ relative to the base pose at the start of the action chunk, and a chunk is a sequence of such waypoints over the horizon. This choice addresses a property of base velocity labels. Mobile bases run under conservative velocity limits with brief acceleration phases, so recorded linear and angular velocities lie mostly either near zero or at the velocity limits. As a result, the velocity labels follow a strongly bimodal distribution, much like the binary open and close states of a gripper. Neural networks learn continuous, smoothly varying targets more easily than such bimodal ones, which are known to be hard to fit with a continuous action model~\citep{chi2023diffusion}. Converting base motion into pose increments turns the bimodal velocity labels into continuous values, because the increments accumulate the velocities over the chunk and vary smoothly along an approach trajectory. Pose increments also remain easy to execute, because a low-level controller converts them into target velocities for the base. The ablation in Sec.~\ref{sec:uniwam-ablations} confirms this design, especially for long approaches.

\textbf{Manipulation actions.} Each arm is represented by 10 dimensions, namely a 3D end-effector position, a 6D rotation, and a 1D gripper state, so the dual-arm action has $(3+6+1)\times2=20$ dimensions. The 6D rotation consists of the first two columns of the rotation matrix, and the full matrix is recovered by Gram--Schmidt orthogonalization~\citep{zhou2019continuity}. Unlike Euler angles and quaternions, this representation is continuous, so nearby rotations have nearby targets. Positions and rotations are expressed in the current main camera frame, which makes visually similar motions numerically similar across embodiments~\citep{yuan2026qwen-robotmanip}. For single-arm robots, the absent arm is zero-padded and masked from the loss.

\subsubsection{Mixed-Stream Batch Packing and Attention}
\label{app:attention-mask}

Encoded examples of both streams are stacked along the batch dimension,
\begin{equation}
X^{\mathrm{pack}}=\operatorname{Concat}_{B}\big(X^{\mathrm{manip}},X^{\mathrm{nav}}\big)\in\mathbb R^{(B_{\mathrm{manip}}+B_{\mathrm{nav}})\times L\times D},
\label{eq:uniwam-batch-packing}
\end{equation}
where $B_{\mathrm{manip}}$ and $B_{\mathrm{nav}}$ need not be equal. Visual tokens and conditioning are packed in the same order, and attention never crosses batch rows. A navigation row therefore conditions only on navigation views and actions, and a manipulation row only on manipulation ones, while both update the same DiT parameters. The two streams therefore interact only through the weights they share.

Within a row, the video DiT reads visual tokens, while language and robot state enter through cross-attention. Current-observation tokens cannot read future frames, future-frame tokens may read the full visual sequence, and action tokens cannot read future video tokens. Let $\mathcal V_t$ denote the current visual tokens and $\mathcal T^s$ the learnable tokens of stream $s$. The action token $a_{t,j}^s$ attends to
\begin{equation}
\mathcal N\big(a_{t,j}^s\big)=\mathcal V_t\,\cup\,\mathcal T^s\,\cup\,\{a_{t,k}^s: k\leq j\},
\label{eq:uniwam-mask}
\end{equation}
so each action reads the current visual context, the learnable tokens, itself, and earlier actions, but not later actions. Learnable tokens read the current visual context and one another, but not action tokens. Under prefix-conditioned asynchronous inference, the first $P$ actions of a chunk are the clean actions already committed for execution. The causal mask of Eq.~\ref{eq:uniwam-mask} lets the later actions condition on this clean prefix without leaking information from later noisy actions backward, so each new chunk continues the executed motion consistently. At deployment, both streams run continuously in one batch without switching model parameters or invoking a separate policy.

\subsubsection{Independent Stream Sampling}
\label{app:sampling-details}

In a complete mobile-manipulation trajectory, the arm is mostly idle during navigation and the base is mostly idle during manipulation. We therefore split each mobile-manipulation trajectory by rule into navigation windows and manipulation windows, one set for each stream. Within each stream, a second rule marks every window as active or idle, depending on whether the controlled part of the robot moves. Each window then becomes a separate example of its stream. A mobile-manipulation trajectory thus contributes to both streams, whereas navigation-only data such as MAP-Data and manipulation-only data contribute only to their own stream. Examples are drawn from
\begin{equation}
q(d,i)=\frac{\alpha_d}{\sum_{d'}\alpha_{d'}}\frac{w_{d,i}}{\sum_j w_{d,j}},
\label{eq:uniwam-sampling}
\end{equation}
where $\alpha_d$ is the weight of source $d$ and $w_{d,i}$ the weight of its $i$-th window. Normalizing within each source decouples its sampling mass from its size, and the window weights set the proportions of navigation and manipulation windows and of active and idle windows. In this way, idle windows are retained at a controlled proportion instead of dominating training. Duplicate or temporally adjacent windows are rejected within a batch, and windows from the same episode must have start frames separated by at least $\delta$ frames. The selected examples are grouped by stream and packed with Eq.~\ref{eq:uniwam-batch-packing}. The stream-specific batch sizes may differ, and navigation and manipulation rows need not originate from the same trajectory.

\subsubsection{Image-Plane Auxiliary Supervision}
\label{app:eef-labels}

Both streams predict spatial actions from images. To ground these actions in the image, we append image-plane targets to each action vector,
\begin{equation}
a_{t,i}^s=\big[a_{t,i}^{s,\mathrm{act}};\,a_{t,i}^{s,\mathrm{aux}}\big],
\label{eq:uniwam-aux}
\end{equation}
and denoise them jointly with the actions, as shown in the output row of Fig.~\ref{fig:uniwam}. For navigation, the targets are the target bounding box and the projected MAP position of Eq.~\ref{eq:map-image-projection}, which indicate where the target is and where the base should stop at every step of the approach. For manipulation, the targets are the two end-effector trajectories projected into the main camera view, together with visibility flags. Predicting these tracks jointly with the camera-frame end-effector poses explicitly links the camera-frame actions to their image-plane locations in the main view. The projected MAP and end-effector positions are image-plane counterparts of the navigation goal and the manipulation action, which keeps the auxiliary targets geometrically consistent with the actions they accompany.

The end-effector tracks are generated from recorded robot states and camera calibration in three steps. Algorithm~\ref{alg:calibrated-eef-geometry} projects each end effector into the main view, Algorithm~\ref{alg:dino-eef-visibility} decides in which frames the projected point is visible, and Algorithm~\ref{alg:eef-labeling} runs both steps on every episode and validates the output.

\textbf{Calibrated geometry.} For each arm, Algorithm~\ref{alg:calibrated-eef-geometry} decodes the end-effector pose in the native base frame of that arm, maps it into a common reference frame with the per-arm transform $M^i_{\mathrm{ref}\leftarrow b_i}$, and then into the main camera frame with $E_{c\leftarrow\mathrm{ref}}$. Two points are tracked per arm, namely the flange of the sixth joint with zero offset and a point on the gripper with offset $\mathbf o_{\mathrm{vis}}$, which is the part seen in the image. Both points are projected with the pinhole model $\pi_K$. A projected gripper point is geometrically valid if it lies in front of the camera, falls inside the image, and belongs to the manipulation interval given by the operation mask $m^{\mathrm{op}}$. The calibration is checked before projection, and a record of its inputs is stored with the labels, so every label can be traced to its calibration.

\textbf{Appearance-gated visibility.} A geometrically valid point may still be hidden, for example by the other arm or by an object. Algorithm~\ref{alg:dino-eef-visibility} thus checks the image content at the projected point. It encodes every $k$-th frame with DINOv2~\citep{oquab2023dinov2}, takes the patch token at the projected gripper point, and scores it by its highest cosine similarity to positive end-effector prototypes minus its highest similarity to negative background prototypes. The scores are interpolated only within contiguous valid segments, thresholded with hysteresis so that the visibility does not flicker near the threshold, and cleaned by filling short gaps and removing short runs. The output keeps the coordinates only in visible frames and labels each frame as visible, valid but occluded, or invalid for the auxiliary loss.

\textbf{Label generation.} Algorithm~\ref{alg:eef-labeling} runs the two steps on every episode in parallel. It resolves the calibration of each episode from a registry, writes the labels together with a per-episode report, and validates all outputs against the dataset before training. Coordinates are normalized to the annotation canvas, and only visible points enter the auxiliary loss.

\begin{algorithm}[t]
\caption{Calibrated main-view end-effector geometry}
\label{alg:calibrated-eef-geometry}
\begin{algorithmic}[1]
\Require states $S_{0:T-1}$; main-view video $V$; intrinsics $K$; reference-to-camera transform $E_{c\leftarrow\mathrm{ref}}$; per-arm base-to-reference transforms $\{M^i_{\mathrm{ref}\leftarrow b_i}\}_{i\in\mathcal A}$; operation mask $m^{\mathrm{op}}$; point offsets $\mathbf o_{\mathrm{J6}}=\mathbf 0$ and $\mathbf o_{\mathrm{vis}}$; arms $\mathcal A=\{\mathrm{left},\mathrm{right}\}$
\Ensure flange and gripper image points $P^{\mathrm{J6}}_{0:T-1}$ and $P^{\mathrm{vis}}_{0:T-1}$; geometric validity mask $m^{\mathrm{geo}}_{0:T-1}$; provenance record $\mathcal R$
\Statex \textbf{Notation:} $h(\mathbf y)=[\mathbf y^{\top},1]^{\top}$; $\operatorname{xyz}(\cdot)$ drops the homogeneous coordinate; $\pi_K([X,Y,Z]^{\top})=[f_xX/Z+c_x,\ f_yY/Z+c_y]^{\top}$
\State $\mathcal R\gets\textsc{ValidateCalibration}(K,E_{c\leftarrow\mathrm{ref}},\{M^i_{\mathrm{ref}\leftarrow b_i}\})$
\For{$t=0,\ldots,T-1$ and $i\in\mathcal A$}
\State $(\mathbf x^i_t,R^i_t)\gets\textsc{DecodeState}(S_t,i)$ \Comment{pose in the arm base frame}
\For{$\rho\in\{\mathrm{J6},\mathrm{vis}\}$}
\State $\mathbf c^{i,\rho}_t\gets\operatorname{xyz}\big(E_{c\leftarrow\mathrm{ref}}\,M^i_{\mathrm{ref}\leftarrow b_i}\,h(\mathbf x^i_t+R^i_t\mathbf o_\rho)\big)$ \Comment{point in the camera frame}
\State $P^{i,\rho}_t\gets\pi_K(\mathbf c^{i,\rho}_t)$
\EndFor
\State $m^{\mathrm{geo},i}_t\gets\mathbf 1[(\mathbf c^{i,\mathrm{vis}}_t)_z>0]\cdot\mathbf 1[P^{i,\mathrm{vis}}_t\in[0,W_{\mathrm{img}})\times[0,H_{\mathrm{img}})]\cdot m^{\mathrm{op}}_t$
\EndFor
\State $\mathcal R\gets\mathcal R\cup\textsc{HashInputs}(S,V,K,E_{c\leftarrow\mathrm{ref}},\{M^i_{\mathrm{ref}\leftarrow b_i}\})$
\State \Return $P^{\mathrm{J6}},P^{\mathrm{vis}},m^{\mathrm{geo}},\mathcal R$
\end{algorithmic}
\end{algorithm}

\begin{algorithm}[t]
\caption{Appearance-gated visibility of the end-effector track}
\label{alg:dino-eef-visibility}
\begin{algorithmic}[1]
\Require main-view video $V$; gripper points $P^{\mathrm{vis}}$; geometric validity mask $m^{\mathrm{geo}}$; DINOv2 encoder $\Phi$; positive and negative prototypes $\mathcal Q^{+},\mathcal Q^{-}$; stride $k$; thresholds $\eta_{\mathrm{in}}>\eta_{\mathrm{out}}$; minimum gap and run lengths $n_{\mathrm{gap}},n_{\mathrm{run}}$; arms $\mathcal A$
\Ensure visibility mask $z_{0:T-1}$; visible-only coordinates $\widetilde P_{0:T-1}$; quality labels $q_{0:T-1}$
\State $U^{+}\gets\textsc{EncodePrototypes}(\Phi,\mathcal Q^{+})$, \ $U^{-}\gets\textsc{EncodePrototypes}(\Phi,\mathcal Q^{-})$
\For{$t\in\{0,k,2k,\ldots\}$ with $t<T$}
\State $F_t\gets\Phi(V_t)$
\For{$i\in\mathcal A$}
\If{$m^{\mathrm{geo},i}_t=1$}
\State $\mathbf f\gets\textsc{PatchToken}(F_t,P^{i,\mathrm{vis}}_t)$ \Comment{token at the projected point}
\State $\mu^i_t\gets\max_{\mathbf u\in U^{+}}\cos(\mathbf f,\mathbf u)-\max_{\mathbf u\in U^{-}}\cos(\mathbf f,\mathbf u)$
\Else
\State $\mu^i_t\gets\mathrm{NaN}$
\EndIf
\EndFor
\EndFor
\State $\bar\mu\gets\textsc{InterpolateWithinSegments}(\mu,m^{\mathrm{geo}})$
\State $z\gets\textsc{Hysteresis}(\bar\mu,\eta_{\mathrm{in}},\eta_{\mathrm{out}})\cdot m^{\mathrm{geo}}$
\State $z\gets\textsc{Morphology}(z,m^{\mathrm{geo}},n_{\mathrm{gap}},n_{\mathrm{run}})$ \Comment{fill short gaps, remove short runs}
\For{$t=0,\ldots,T-1$ and $i\in\mathcal A$}
\State $\widetilde P^i_t\gets P^{i,\mathrm{vis}}_t$ if $z^i_t=1$, else $\mathrm{NaN}$
\State $q^i_t\gets 1$ if $z^i_t=1$; $2$ if $m^{\mathrm{geo},i}_t=1$ and $z^i_t=0$; $0$ otherwise \Comment{visible, occluded, invalid}
\EndFor
\State \Return $z,\widetilde P,q$
\end{algorithmic}
\end{algorithm}

\begin{algorithm}[t]
\caption{End-effector label generation for a dataset}
\label{alg:eef-labeling}
\begin{algorithmic}[1]
\Require episodes $\mathcal D$; calibration registry $\mathcal C$; point offsets $\mathbf o_{\mathrm{J6}},\mathbf o_{\mathrm{vis}}$; visibility parameters $\Theta_{\mathrm{vis}}=(\Phi,\mathcal Q^{+},\mathcal Q^{-},k,\eta_{\mathrm{in}},\eta_{\mathrm{out}},n_{\mathrm{gap}},n_{\mathrm{run}})$
\Ensure per-episode labels and reports; validated manifest $\mathcal M$
\ForAll{episodes $e\in\mathcal D$ in parallel}
\State $(S^e,V^e)\gets\textsc{ReadEpisode}(e)$
\State $(K^e,E^e,\{M^{i,e}\},m^{\mathrm{op},e})\gets\textsc{ResolveCalibration}(e,\mathcal C)$
\State $(P^{\mathrm{J6},e},P^{\mathrm{vis},e},m^{\mathrm{geo},e},\mathcal R^e)\gets$ Algorithm~\ref{alg:calibrated-eef-geometry}
\State $(z^e,\widetilde P^e,q^e)\gets$ Algorithm~\ref{alg:dino-eef-visibility} with $\Theta_{\mathrm{vis}}$
\State $\textsc{WriteLabels}(e,P^{\mathrm{J6},e},P^{\mathrm{vis},e},\widetilde P^e,m^{\mathrm{geo},e},z^e,q^e,\mathcal R^e)$ \Comment{labels and report}
\EndFor
\State $\mathcal M\gets\textsc{ValidateAll}(\mathcal D)$ \Comment{every episode has complete labels}
\State \Return $\mathcal M$
\end{algorithmic}
\end{algorithm}

\subsubsection{Training Objective}
\label{app:uniwam-training-details}

All targets are trained with flow matching~\citep{lipman2023flow}. For a clean video-latent sequence or normalized action chunk $\mathbf x$, noise $\boldsymbol\epsilon\sim\mathcal N(\mathbf 0,\mathbf I)$, and noise level $\sigma\in[0,1]$, the noisy input and the velocity target are
\begin{equation}
\mathbf x_\sigma=(1-\sigma)\mathbf x+\sigma\boldsymbol\epsilon,\qquad\mathbf v^*=\boldsymbol\epsilon-\mathbf x.
\label{eq:uniwam-flow}
\end{equation}
Video and action noise levels are sampled independently for each example as $\sigma=\kappa u/[1+(\kappa-1)u]$ with $u\sim\mathcal U(0,1)$ and $\kappa=5$. The current observation remains clean, and only future video latents are supervised. The action and auxiliary dimensions of one action vector share a noise level. For prefix-conditioned training, positions $1\leq t\leq P$ are replaced by clean committed actions and excluded from supervision. With probability $p_{\mathrm{pre}}$, $P$ is drawn uniformly from $\{P_{\min},\ldots,P_{\max}\}$, and otherwise $P=0$, which matches the prefix lengths used at deployment.

For each supervised group $g$, let $\mathcal F_{b,t}^{g}$ contain the valid features at position $t$ of example $b$, and let $\mathcal I_b^g$ contain the non-prefix, non-padded positions with at least one valid feature. The per-example loss is
\begin{equation}
\ell_b^g=\frac{\omega(\sigma_b)}{|\mathcal I_b^g|}\sum_{t\in\mathcal I_b^g}\frac{1}{|\mathcal F_{b,t}^{g}|}\sum_{j\in\mathcal F_{b,t}^{g}}\left(\widehat v_{b,t,j}-v^*_{b,t,j}\right)^2,
\label{eq:uniwam-group-loss}
\end{equation}
where $\omega(\sigma_b)$ is a loss weight that depends on the noise level. Each group loss $\mathcal L^g$ is the mean of $\ell_b^g$ over the examples with valid supervision, using counts aggregated across workers. The full objective is
\begin{equation}
\mathcal L=\lambda_{\mathrm{vid}}\mathcal L_{\mathrm{vid}}+\sum_{s\in\{\mathrm{nav},\mathrm{manip}\}}\big(\lambda_{\mathrm{act}}^{s}\mathcal L_{\mathrm{act}}^{s}+\lambda_{\mathrm{aux}}^{s}\mathcal L_{\mathrm{aux}}^{s}\big),
\label{eq:uniwam-objective}
\end{equation}
where $\mathcal L_{\mathrm{vid}}$ supervises future video latents of both streams, $\mathcal L_{\mathrm{act}}^{s}$ the actions, and $\mathcal L_{\mathrm{aux}}^{s}$ the image-plane targets. Because each term is averaged only over valid positions and features, a sample contributes only to the losses of its own stream.

\subsubsection{Implementation and Inference}
\label{app:implementation-details}

\textbf{Network and optimization.} UniWAM uses the Wan2.2-TI2V-5B~\citep{wan2025wan} video backbone, video VAE, and text conditioning. The action DiT has 30 transformer layers and a hidden dimension of $D_a=1024$, compared with $D_v=3072$ for the video DiT. Sampling every fourth image from a 33-frame observation window yields nine RGB frames, which are encoded into three temporal VAE latents. Camera views are arranged into a $384\times320$ canvas with a main view and wrist panels. We optimize with AdamW, an initial learning rate of $10^{-5}$, and cosine decay. The video VAE and the text encoder stay frozen, and all other parameters are trained. Inference uses a deterministic schedule with ten Euler denoising steps and no classifier-free guidance for either stream.

\textbf{Infrastructure acceleration.} In standard PyTorch execution, each transformer operation is dispatched separately from Python. Because the mixed video-action layers operate on relatively short action sequences, the CPU may spend substantial time submitting kernels while the GPU waits for the next launch. This overhead affects every training step and becomes more pronounced during inference, where the action denoising computation is repeated over multiple Euler steps. UniWAM reduces this overhead through separate training and inference paths. During training, \texttt{torch.compile} captures the tensor-only mixed-stream transformer layer as a full graph and reuses the compiled function across layers and optimization steps, while data preparation, loss computation, and parameter updates remain outside the compiled region. During inference, we compile the VAE encoder, the visual-cache prefill, and the cached action-denoising function separately. The current observation is encoded once, and its layerwise video keys and values are reused across the ten Euler steps, so the video branch is not recomputed at every step. For fixed input shapes, reduce-overhead compilation and execution compatible with CUDA graphs further reduce the cost of CPU submission and kernel launches. When both streams are active, navigation and manipulation rows are packed into the same batch and processed by the same compiled functions, which increases the work in each launch and avoids separate model invocations. In our measurements, this implementation reduces two-stream inference latency on one NVIDIA H200 from about 220\,ms to below 60\,ms, and the corresponding training setup exceeds 600 samples per second on 16 NVIDIA H200 GPUs. Asynchronous execution further overlaps the remaining inference time with the execution of committed actions.

\section{Experiments}
\label{sec:uniwam-experiments}

We evaluate UniWAM on MAP-Bench and on 24 real-robot tasks. The experiments answer four questions. First, does UniWAM dock more accurately than existing policies, and how much does MAP-Data help? Second, does this advantage carry over to real robots and to complete mobile-manipulation tasks? Third, which part of the gain comes from the mixed-stream architecture and which from independent sampling? Fourth, how does each design component contribute?

\subsection{Experimental Setup}
\label{sec:uniwam-setup}

\textbf{Tasks and platforms.} MAP-Bench evaluates MAP navigation from held-out initial poses in the three distance ranges of Sec.~\ref{sec:map}. The real-robot evaluation covers 13 MAP Navigation, 4 Mobile Manipulation, and 7 Manipulation tasks on four embodiments, namely AgileX Cobot Magic, AgiBot G2, dual-arm Franka Research 3, and an AgileX single-arm platform. The 13 MAP Navigation tasks run on both the Cobot Magic and the AgiBot G2. The four Mobile Manipulation tasks and Manipulation task 5 run on the Cobot Magic, Manipulation tasks 1 to 3 and 7 on the dual Franka, and Manipulation tasks 4 and 6 on the AgileX single arm, so every embodiment contributes to the evaluation.

\begin{figure}[t]
\centering
\includegraphics[width=0.55\linewidth]{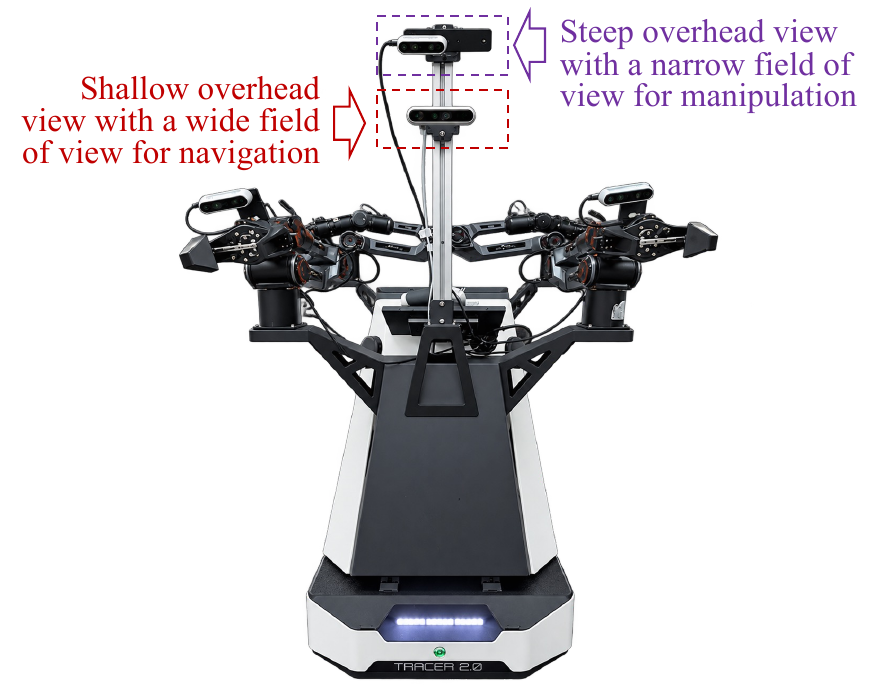}
\caption{Mobile manipulation platform. A shallow-angle head camera with a wide field of view provides the navigation-stream input, and a steep-angle head camera with a narrow field of view, together with the wrist cameras, provides the manipulation-stream input. In Mobile Manipulation, the navigation stream also receives the wrist cameras. The two inputs form the navigation and manipulation rows of a mixed-stream batch, which UniWAM processes in one forward pass.}
\label{fig:platform}
\end{figure}

Fig.~\ref{fig:platform} shows the mobile platform, whose two head cameras realize the stream-specific observations of UniWAM. The wide-view camera captures distant targets and free space for navigation, and the downward-facing camera captures the workspace for manipulation. The navigation stream observes the wide-view head camera, and the manipulation stream observes the downward-facing head camera and the two wrist cameras. In Mobile Manipulation, the navigation stream also observes the wrist cameras to track the arms during the approach.

\textbf{UniWAM variants.} Both variants start from Wan2.2-TI2V-5B~\citep{wan2025wan} without embodied pretraining. \textbf{UniWAM-Base} uses only the data available to the baselines and is the matched comparison. It is trained on the 35K-trajectory training split on MAP-Bench and on the demonstrations of the 24 evaluation tasks on real robots. \textbf{UniWAM-Scaled} keeps the architecture and adds data through independent stream sampling, drawing large-scale and task data at a 7:3 ratio. On MAP-Bench, it adds MAP-Data under the same training budget, so its gain isolates the value of MAP-Data. On real robots, it adds MAP-Data to the navigation stream, sampling 15 trajectories per navigation target, and adds 402 private manipulation tasks with 30 demonstrations each to the manipulation stream. These tasks are disjoint from the evaluation tasks. In both variants, windows of mobile-manipulation trajectories in which the arm is idle during base motion are sampled at 20\% of their original probability, and windows in which the base is idle during manipulation at 15\%. Each variant is a single model for all tasks and embodiments, without task-specific fine-tuning.

\textbf{MAP-Bench baselines.} We compare with the VLA models $\pi_{0.5}$~\citep{black2025pi05} and GR00T N1.7~\citep{bjorck2025gr00t}, the VLN models Uni-NaVid~\citep{zhang2024uni} and InternVLA-N1~\citep{internvla2025}, and the world-action model FastWAM~\citep{yuan2026fastwam}. All models, including UniWAM, are trained on the MAP-Bench training split for 40,000 steps on 16 NVIDIA H200 GPUs. All models observe only the forward RGB camera, and each keeps its native action interface, into which the navigation labels are converted. Only UniWAM receives the bounding-box and MAP auxiliary targets. $\pi_{0.5}$ and GR00T N1.7 are fine-tuned with all parameters from their public checkpoints. InternVLA-N1 starts from the released DualVLN checkpoint~\citep{wei2026ground} and, as in its official recipe, keeps System 2 frozen and trains only the System 1 trajectory generator, using the current frame and up to four history frames. Uni-NaVid is fine-tuned with all parameters from the Uni-NaVid-7B checkpoint, using the current observation and up to three earlier observations sampled every 0.5\,s. Because its native outputs are discrete forward and turning primitives, its labels are the primitive sequences that best fit the future trajectory, found by dynamic programming. FastWAM is the baseline closest to UniWAM in formulation. It starts from Wan2.2-TI2V-5B and shares the data, visual inputs, action targets, horizon, optimizer, and schedule of UniWAM-Base, but follows its own official architecture and does not use the stream-specific interfaces, the causal action mask, or the image-plane auxiliary supervision. Apart from the shared data and budget, each baseline follows its official fine-tuning recipe.

\textbf{Real-robot baselines.} We compare with $\pi_{0.5}$ and GR00T N1.7, which start from their public checkpoints, keep their native action interfaces, and update all parameters. All methods are trained on the same demonstrations of the 24 evaluation tasks with the same instructions. Each MAP Navigation task has 50 demonstrations on each of its two embodiments, and each Mobile Manipulation or Manipulation task has about 200, with 400 for selected challenging tasks such as six-cup stacking. All methods observe the same cameras. In MAP Navigation, every method observes only the wide-view navigation head camera. In Mobile Manipulation, all methods observe the navigation head camera, the manipulation head camera, and the two wrist cameras. UniWAM assigns them to its streams, whereas $\pi_{0.5}$ and GR00T N1.7 receive all four views as a single multi-view observation. For this four-camera setting, GR00T N1.7 uses a bimanual embodiment configuration similar to that of ALOHA, with one video key per camera. For MAP Navigation, each baseline is trained as one model on all 13 tasks for each of the Cobot Magic and the AgiBot G2, because these tasks share the action space and differ mainly in the target. For the other tasks, each baseline is fine-tuned separately for each embodiment. We also co-trained each baseline on all 24 tasks, as for UniWAM, but this performed worse than per-embodiment training for both $\pi_{0.5}$ and GR00T N1.7, which we attribute to their large-scale robot pretraining. We therefore report their per-embodiment models, each evaluated at the checkpoint that performed best in our real-robot tests. Table~\ref{tab:new-controls} summarizes the training settings of all real-robot models, including initialization, data, and compute. Sharing the data, cameras, and evaluation protocol across all methods keeps the comparison fair.

\begin{table}[t]
\caption{Real-robot training settings. Steps and GPUs are given per model. Each baseline trains one MAP Navigation model per mobile robot and one model for the remaining tasks of each embodiment, whereas each UniWAM variant is one model for all tasks.}
\label{tab:new-controls}
\centering
\small
\setlength{\tabcolsep}{4pt}
\begin{tabular}{@{}lcccc@{}}
\toprule
 & $\pi_{0.5}$ & GR00T N1.7 & UniWAM-Base & UniWAM-Scaled \\
\midrule
Initialization & $\pi_{0.5}$ base checkpoint & GR00T N1.7-3B checkpoint & Wan2.2-TI2V-5B & Wan2.2-TI2V-5B \\
Robot pretraining & yes & yes & none & none \\
Frozen modules & none & none & VAE, text encoder & VAE, text encoder \\
Task demonstrations & 24 tasks & 24 tasks & 24 tasks & 24 tasks \\
Additional data & none & none & none & MAP-Data and 402 tasks \\
Image-plane auxiliary outputs & \xmark & \xmark & \cmark & \cmark \\
Steps per model & 50K & 50K & 20K & 30K \\
GPUs per model & 8 & 8 & 32 & 64 \\
\bottomrule
\end{tabular}
\end{table}

\textbf{Real-robot evaluation protocol.} The reference MAP of each task is the parking pose marked during data collection. In each MAP Navigation trial, the robot starts at this pose and is moved to a random initial pose 1 to 5\,m away with a heading offset within $\pm45^\circ$. The onboard state estimator records both motions, so the reached pose is scored against the same reference. The robot runs until it stops under the rule of Sec.~\ref{sec:map}. Each Mobile Manipulation trial consists of navigation followed by manipulation. When the base stops, $E_{\mathrm{pos}}$ and $E_{\mathrm{head}}$ are computed against the reference MAP, and the robot then executes the manipulation stage. At test time, UniWAM activates both streams and packs them into one batch throughout each trial. The policies still predict small but nonzero base motions during manipulation, which can disturb precise placements. Therefore, for all methods, a predicted base chunk is not executed when its displacement stays below 0.05\,m and 0.05\,rad, analogous to the stopping rule of MAP-Bench. Each Manipulation trial starts from a fixed working pose, and a task-specific predicate decides success. Each task is evaluated over 40 trials on each embodiment it runs on, and MAP Navigation results are averaged over the Cobot Magic and the AgiBot G2. The initial configurations of these trials, including the robot pose and the object placements, are recorded once and reproduced for every method. All methods share task definitions, instructions, success criteria, and termination rules.

\textbf{Metrics.} MAP Navigation is measured by $E_{\mathrm{pos}}$ and $E_{\mathrm{head}}$ of Eq.~\ref{eq:map-metrics}. On real robots, the reached pose is obtained from the onboard state estimator. Mobile Manipulation reports the same errors when the base stops, together with the success rate (SR) of the subsequent manipulation. Manipulation reports SR. Real-robot results are unweighted task means.

\subsection{MAP-Bench Results}
\label{sec:mapbench-results}

\begin{table}[t]
\caption{MAP-Bench results over the evaluation split. Bold and underline mark the best and second-best results.}
\label{tab:mapbench-baselines}
\centering
\small
\setlength{\tabcolsep}{5pt}
\begin{tabular}{@{}lrrr r||rrr r@{}}
\toprule
& \multicolumn{3}{c}{\textbf{$E_{\mathrm{pos}}$ (cm) $\downarrow$}} & \multicolumn{1}{c||}{\multirow{2}{*}{\textbf{Overall}}} & \multicolumn{3}{c}{\textbf{$E_{\mathrm{head}}$ ($^\circ$) $\downarrow$}} & \multirow{2}{*}{\textbf{Overall}} \\
\cmidrule(lr){2-4}\cmidrule(lr){6-8}
\textbf{Model} & \textbf{Short} & \textbf{Mid} & \textbf{Long} & & \textbf{Short} & \textbf{Mid} & \textbf{Long} & \\
\midrule
\rowcolor{blue!10}
\multicolumn{9}{@{}l}{\textit{Vision-language-action models (VLA)}} \\
GR00T N1.7~\citep{bjorck2025gr00t} & 9.48 & 10.8 & 22.0 & 14.1 & 5.66 & 5.42 & 7.46 & 6.18 \\
$\pi_{0.5}$~\citep{black2025pi05} & 7.98 & 9.85 & 50.8 & 22.9 & 3.26 & 3.51 & 5.55 & 4.11 \\
\midrule
\rowcolor{blue!10}
\multicolumn{9}{@{}l}{\textit{Vision-language navigation models (VLN)}} \\
Uni-NaVid~\citep{zhang2024uni} & 17.3 & 40.0 & 87.2 & 48.2 & 7.98 & 15.0 & 23.4 & 15.5 \\
InternVLA-N1~\citep{internvla2025} & 17.3 & 39.7 & 130.0 & 62.3 & 3.72 & 11.0 & 27.6 & 14.1 \\
\midrule
\rowcolor{blue!10}
\multicolumn{9}{@{}l}{\textit{World-action models (WAM)}} \\
FastWAM~\citep{yuan2026fastwam} & 9.24 & 11.0 & 31.1 & 17.1 & 3.44 & 3.95 & 6.66 & 4.68 \\
\textbf{UniWAM-Base} & \underline{7.71} & \underline{8.85} & \underline{21.0} & \underline{12.5} & \underline{2.37} & \underline{3.06} & \underline{4.29} & \underline{3.24} \\
\textbf{UniWAM-Scaled} & \textbf{7.48} & \textbf{7.86} & \textbf{14.2} & \textbf{9.85} & \textbf{1.82} & \textbf{2.08} & \textbf{3.01} & \textbf{2.30} \\
\bottomrule
\end{tabular}
\end{table}

Table~\ref{tab:mapbench-baselines} summarizes MAP-Bench performance across models and initial distances.

\textbf{Overall comparison.} UniWAM-Scaled achieves the lowest errors in every distance range, reaching 9.85\,cm and $2.30^\circ$ overall. These errors are 30.1\% and 44.0\% lower than those of the strongest external baselines, GR00T N1.7 at 14.1\,cm and $\pi_{0.5}$ at $4.11^\circ$, respectively. Without MAP-Data, UniWAM-Base already reduces these errors by 11.3\% and 21.2\%, which shows that the architecture alone improves docking under identical data.

\textbf{Semantic versus manipulation-ready navigation.} Although trained on the same split, the VLN models retain large terminal errors, with Overall position errors of 48.2 and 62.3\,cm and long-range errors of 87.2 and 130.0\,cm. Both are designed for semantic navigation, and their official recipes keep the planning module of InternVLA-N1 frozen and restrict Uni-NaVid to discrete motion primitives. The VLA and WAM baselines are more accurate but degrade with distance. For example, the position error of $\pi_{0.5}$ increases from 7.98\,cm at short range to 50.8\,cm at long range.

\textbf{Long-range docking.} Relative to the best baseline in each range, UniWAM-Scaled reduces position error by 6.3\% at short range and 35.5\% at long range. Its position error grows by a factor of 1.9 from short to long range, compared with 2.3 for GR00T N1.7, 3.4 for FastWAM, and 6.4 for $\pi_{0.5}$. Its heading error is also 40.7\% to 45.8\% lower than that of the best baseline in each of the three distance ranges.

\textbf{Effect of MAP-Data.} UniWAM-Base and UniWAM-Scaled share the architecture and training budget, and only UniWAM-Scaled mixes MAP-Data. MAP-Data reduces the Overall position and heading errors by 21.2\% and 29.0\%, and the position gain increases from 3.0\% at short range to 32.4\% at long range. Long approaches thus benefit most from explicit supervision of where to stop, since small heading errors grow over distance.

\textbf{Scaling with MAP-Data.} This experiment asks whether navigation keeps improving as more MAP-Data becomes available. We train UniWAM on MAP-Bench with a random subset of 25\%, 50\%, or 100\% of the MAP-Data targets, sampling 15 trajectories per target and excluding every benchmark target. All other settings follow UniWAM-Scaled, including the architecture, the 7:3 mixture of MAP-Data and the MAP-Bench training split, and 40,000 steps on 16 GPUs. Because the mixture ratio and the number of steps are fixed, the 25\%, 50\%, and 100\% runs see the same number of MAP-Data samples and differ only in how many distinct targets and scenes these samples come from. Table~\ref{tab:new-scaling} shows that both errors decrease monotonically as more of MAP-Data is used. The gain concentrates at long range, where the position error falls from 21.0 to 14.2\,cm, whereas the short-range error changes by only 0.23\,cm. The gain diminishes with more data but remains positive, and the last doubling still reduces the Overall position error by 4.4\%.

To test whether other models benefit equally, we also train $\pi_{0.5}$ with the full MAP-Data under the same mixture and budget. Its Overall errors fall by only 5.7\% and 1.0\%, compared with 21.2\% and 29.0\% for UniWAM, and its long-range position error remains at 47.8\,cm, compared with 14.2\,cm for UniWAM-Scaled. $\pi_{0.5}$ gains mainly at short range, where it reaches 7.42\,cm, on par with UniWAM-Scaled. Large-scale MAP supervision thus helps most when navigation is learned as its own stream with image-plane targets, as in UniWAM.

\begin{table}[t]
\caption{MAP-Bench errors when training with a fraction of the MAP-Data targets under a fixed compute budget. The second column is the fraction of MAP-Data targets from which the MAP-Data samples are drawn. Whenever MAP-Data is used, it is mixed with the MAP-Bench training split at a fixed 7:3 ratio, so all runs with MAP-Data see the same number of MAP-Data samples. 0\% denotes training on the MAP-Bench split alone. UniWAM at 0\% and 100\% are UniWAM-Base and UniWAM-Scaled, and $\pi_{0.5}$ uses the same data and budget. Position errors are in centimeters and heading errors in degrees.}
\label{tab:new-scaling}
\centering
\small
\setlength{\tabcolsep}{6pt}
\begin{tabular}{@{}lcccccc@{}}
\toprule
Model & MAP-Data targets & $E_{\mathrm{pos}}$ Short & $E_{\mathrm{pos}}$ Mid & $E_{\mathrm{pos}}$ Long & $E_{\mathrm{pos}}$ Overall & $E_{\mathrm{head}}$ Overall \\
\midrule
UniWAM-Base & 0\% & 7.71 & 8.85 & 21.0 & 12.5 & 3.24 \\
UniWAM & 25\% & 7.63 & 8.52 & 18.1 & 11.4 & 2.81 \\
UniWAM & 50\% & 7.52 & 8.03 & 15.3 & 10.3 & 2.43 \\
UniWAM-Scaled & 100\% & 7.48 & 7.86 & 14.2 & 9.85 & 2.30 \\
\midrule
$\pi_{0.5}$ & 0\% & 7.98 & 9.85 & 50.8 & 22.9 & 4.11 \\
$\pi_{0.5}$ & 100\% & 7.42 & 9.45 & 47.8 & 21.6 & 4.07 \\
\bottomrule
\end{tabular}
\end{table}

\subsection{Real-Robot Results}
\label{sec:real-world-results}

\begin{table}[t]
\caption{Real-robot results averaged over the tasks in each category. Bold and underline mark the best and second-best results.}
\label{tab:real-world-eval}
\centering
\small
\setlength{\tabcolsep}{5pt}
\begin{tabular}{@{}lcc||ccc||c@{}}
\toprule
& \multicolumn{2}{c||}{\textbf{MAP Navigation}} & \multicolumn{3}{c||}{\textbf{Mobile Manipulation}} & \textbf{Manipulation} \\
\cmidrule(lr){2-3}\cmidrule(lr){4-6}\cmidrule(lr){7-7}
\textbf{Model} & \textbf{$E_{\mathrm{pos}}$ (cm) $\downarrow$} & \textbf{$E_{\mathrm{head}}$ ($^\circ$) $\downarrow$} & \textbf{$E_{\mathrm{pos}}$ (cm) $\downarrow$} & \textbf{$E_{\mathrm{head}}$ ($^\circ$) $\downarrow$} & \textbf{SR (\%) $\uparrow$} & \textbf{SR (\%) $\uparrow$} \\
\midrule
$\pi_{0.5}$~\citep{black2025pi05} & 23.3 & 5.15 & 26.2 & 5.72 & 62.5 & \textbf{71.4} \\
GR00T N1.7~\citep{bjorck2025gr00t} & 18.1 & 6.55 & 20.4 & 7.20 & 39.4 & 36.8 \\
\textbf{UniWAM-Base} & \underline{15.1} & \underline{4.91} & \underline{17.1} & \underline{5.30} & \underline{63.1} & 44.3 \\
\textbf{UniWAM-Scaled} & \textbf{11.7} & \textbf{4.25} & \textbf{13.2} & \textbf{4.56} & \textbf{80.0} & \underline{68.9} \\
\bottomrule
\end{tabular}
\end{table}

\begin{figure}[p]
\centering
\includegraphics[width=\textwidth,height=0.88\textheight,keepaspectratio]{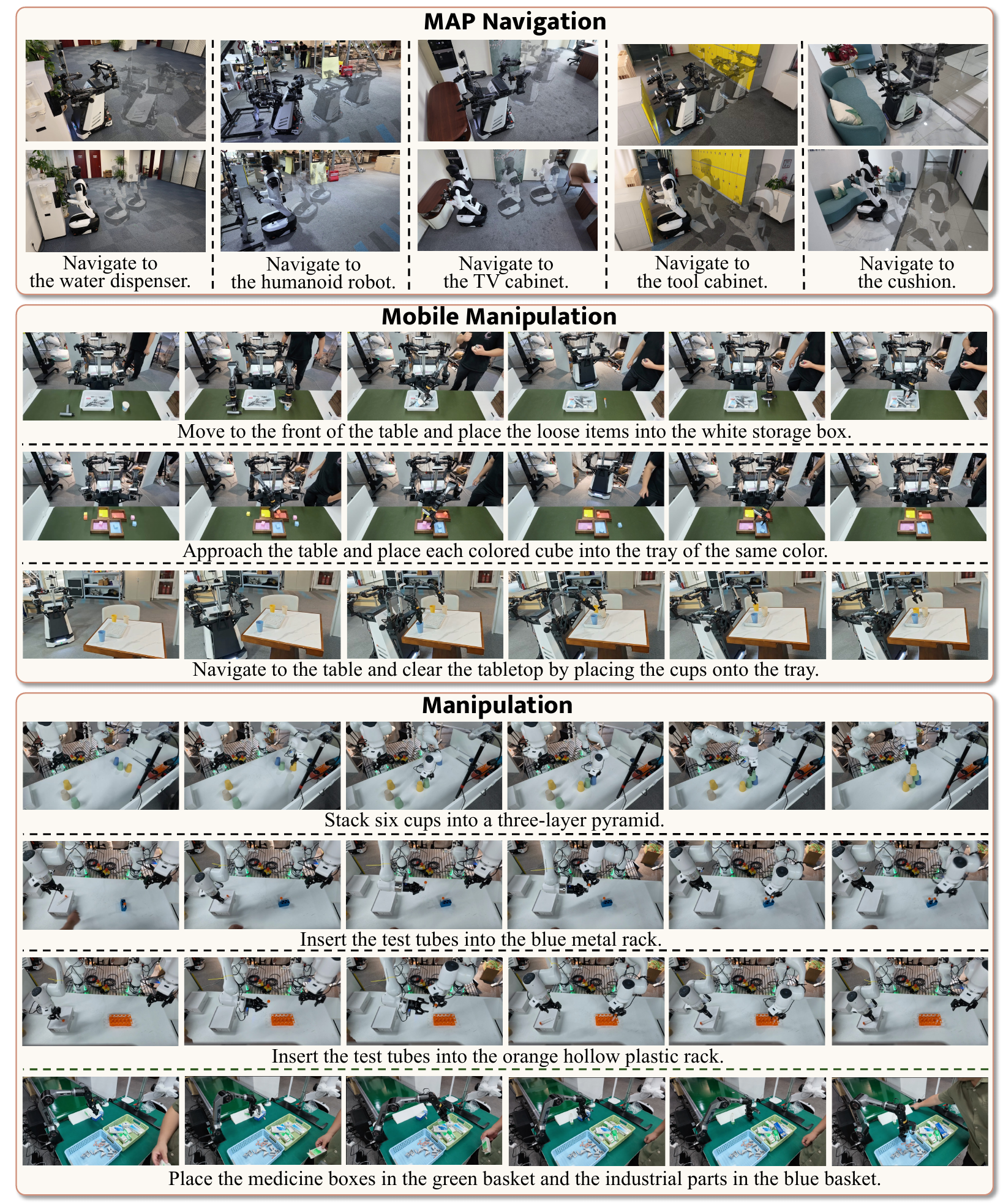}
\caption{Real-robot executions of UniWAM. A single UniWAM model performs MAP Navigation, Mobile Manipulation, and Manipulation across four embodiments. The MAP Navigation examples dock at targets of different categories and sizes, the Mobile Manipulation examples show the transition from approach to manipulation, and the Manipulation examples show object sorting, test-tube insertion, and cup stacking.}
\label{fig:real-world-qualitative}
\end{figure}

Table~\ref{tab:real-world-eval} summarizes the 24 real-robot tasks, and Fig.~\ref{fig:real-world-qualitative} in Sec.~\ref{sec:qualitative-results} shows representative executions. Among the evaluated models, UniWAM-Scaled achieves the lowest position and heading errors in MAP Navigation and Mobile Manipulation, the highest Mobile Manipulation success rate, and the second-highest Manipulation success rate.

\textbf{MAP navigation.} The two baselines trade off position and heading accuracy. GR00T N1.7 has the lower position error, 18.1\,cm compared with 23.3\,cm for $\pi_{0.5}$, whereas $\pi_{0.5}$ has the lower heading error, $5.15^\circ$ compared with $6.55^\circ$. Trained only on task data, UniWAM-Base outperforms both on both metrics, reaching 15.1\,cm and $4.91^\circ$. UniWAM-Scaled further reduces the errors to 11.7\,cm and $4.25^\circ$, 35.4\% and 17.5\% below the strongest baseline, respectively. It achieves the lowest position error on all 13 tasks and the lowest heading error on 11.

\textbf{Mobile manipulation.} The docking accuracy of UniWAM-Scaled carries over to end-to-end performance. It stops 13.2\,cm and $4.56^\circ$ from the MAP, 35.3\% and 20.3\% below the best baseline, respectively. Its success rate reaches 80.0\%, 17.5 percentage points above $\pi_{0.5}$ at 62.5\%, with non-overlapping 95\% intervals, and it is highest on all four tasks. UniWAM-Base reaches a success rate similar to that of $\pi_{0.5}$, 63.1\% versus 62.5\%, while docking more accurately.

\textbf{Manipulation.} $\pi_{0.5}$ achieves the highest average success at 71.4\%, followed by UniWAM-Scaled at 68.9\%. UniWAM-Scaled leads on the three sorting and placement tasks, while $\pi_{0.5}$ leads by 2.5 to 12.5 percentage points on six-cup stacking, the two tube-insertion tasks, and long-horizon coffee preparation. $\pi_{0.5}$ is pretrained on large-scale robot data, whereas UniWAM starts from a video model and relies on mixed-data training, so the remaining gap is largest on these dexterous or long-horizon tasks. GR00T N1.7 has the lowest success at 36.8\%.

\textbf{Scaling.} Relative to UniWAM-Base, UniWAM-Scaled reduces MAP Navigation errors by 22.5\% and 13.4\% and raises success by 16.9 and 24.6 percentage points on Mobile Manipulation and Manipulation. These gains reflect both more data and more compute, and Sec.~\ref{sec:mapbench-results} isolates the effect of MAP-Data.

\textbf{Per-task results.} Table~\ref{tab:real-world-task-eval} reports the per-task results. In MAP Navigation, UniWAM-Scaled has the lowest position error on all 13 tasks and the lowest heading error on 11. On the sink and the round side table, UniWAM-Base has the lowest heading error. In Mobile Manipulation, UniWAM-Scaled is best on all four tasks in all three metrics. UniWAM-Base exceeds $\pi_{0.5}$ in success on cube sorting and pouring and trails it by 2.5 points on the other two. In Manipulation, UniWAM-Scaled leads on tasks 4 to 6, and $\pi_{0.5}$ leads on tasks 1 to 3 and 7, with the largest gap of 12.5 points on coffee preparation, where GR00T N1.7 fails every trial and $\pi_{0.5}$ reaches 52.5\%.

\begin{table}[t]
\caption{Per-task real-robot results with 40 trials per task on each embodiment. MAP Navigation cells report position error and heading error averaged over the Cobot Magic and the AgiBot G2. Mobile Manipulation cells additionally report the success rate in percent, and Manipulation cells report only the success rate. Avg.\ is the unweighted mean over the tasks in each category. Bold and underline mark the best and second-best values for each metric.}
\label{tab:real-world-task-eval}
\centering
\small
\setlength{\tabcolsep}{2pt}
\renewcommand{\arraystretch}{1.2}
\begin{tabularx}{\textwidth}{@{}r>{\hsize=1.8\hsize\raggedright\arraybackslash}X*{4}{>{\hsize=.8\hsize\centering\arraybackslash}X}@{}}
\toprule
\textbf{ID} & \textbf{Task} & \textbf{$\pi_{0.5}$} & \textbf{GR00T N1.7} & \textbf{UniWAM-Base} & \textbf{UniWAM-Scaled} \\
\midrule
\rowcolor{blue!10}
\multicolumn{6}{@{}l}{\textit{\textbf{MAP Navigation}} \quad $E_{\mathrm{pos}}$ (cm) $\downarrow$ / $E_{\mathrm{head}}$ ($^\circ$) $\downarrow$} \\
1 & Water dispenser & 18.9 / 4.83 & 19.7 / 5.47 & \underline{13.9} / \underline{4.23} & \textbf{10.3} / \textbf{3.86} \\
2 & Humanoid robot & 24.7 / 5.17 & 18.4 / 6.57 & \underline{15.3} / \underline{4.84} & \textbf{12.6} / \textbf{4.47} \\
3 & TV cabinet & 20.2 / 4.56 & 14.8 / 5.74 & \underline{12.1} / \underline{3.76} & \textbf{8.94} / \textbf{3.58} \\
4 & Tool cabinet & 25.3 / \underline{5.63} & 20.2 / 7.18 & \underline{16.5} / 7.27 & \textbf{11.7} / \textbf{4.06} \\
5 & Cushion & 21.8 / 4.69 & 15.6 / 5.93 & \underline{13.2} / \underline{4.34} & \textbf{9.73} / \textbf{3.74} \\
6 & Table & 28.1 / 5.84 & 19.7 / 7.46 & \underline{16.9} / \underline{5.47} & \textbf{13.7} / \textbf{4.86} \\
7 & Sink & 16.9 / 4.27 & 17.3 / 6.34 & \underline{12.7} / \textbf{3.86} & \textbf{10.8} / \underline{4.13} \\
8 & Potted plant & 30.9 / \underline{6.36} & 23.1 / 8.13 & \underline{19.7} / 6.58 & \textbf{14.8} / \textbf{5.24} \\
9 & Wall-side cabinet & 23.7 / 5.14 & 18.2 / 6.77 & \underline{14.8} / \underline{4.79} & \textbf{11.9} / \textbf{4.31} \\
10 & Workstation & 24.9 / 5.46 & 18.8 / 6.86 & \underline{15.9} / \underline{5.24} & \textbf{13.1} / \textbf{4.72} \\
11 & Black side cabinet & 26.4 / \underline{5.91} & 21.1 / 7.57 & \underline{17.1} / 5.96 & \textbf{12.8} / \textbf{4.57} \\
12 & Round side table & 21.2 / 4.76 & 15.1 / 5.84 & \underline{14.5} / \textbf{3.94} & \textbf{11.8} / \underline{4.24} \\
13 & Countertop & 19.5 / 4.35 & 13.9 / 5.24 & \underline{13.4} / \underline{3.53} & \textbf{9.41} / \textbf{3.46} \\
\cmidrule(l){2-6}
& \textit{Avg.} & 23.3 / 5.15 & 18.1 / 6.55 & \underline{15.1} / \underline{4.91} & \textbf{11.7} / \textbf{4.25} \\
\midrule
\rowcolor{blue!10}
\multicolumn{6}{@{}l}{\textit{\textbf{Mobile Manipulation}} \quad $E_{\mathrm{pos}}$ (cm) $\downarrow$ / $E_{\mathrm{head}}$ ($^\circ$) $\downarrow$ / SR (\%) $\uparrow$} \\
1 & Place items in storage box & 25.6/5.67/\underline{65.0} & 19.2/6.91/42.5 & \underline{16.7}/\underline{5.16}/62.5 & \textbf{12.8}/\textbf{4.37}/\textbf{82.5} \\
2 & Sort colored cubes into trays & 23.8/\underline{5.29}/55.0 & 21.3/7.46/35.0 & \underline{18.2}/5.73/\underline{60.0} & \textbf{13.1}/\textbf{4.46}/\textbf{77.5} \\
3 & Clear cups onto tray & 29.1/6.18/\underline{70.0} & 18.7/6.58/47.5 & \underline{15.9}/\underline{4.87}/67.5 & \textbf{13.8}/\textbf{4.83}/\textbf{85.0} \\
4 & Pour water into cup & 26.3/5.74/60.0 & 22.4/7.83/32.5 & \underline{17.6}/\underline{5.42}/\underline{62.5} & \textbf{13.1}/\textbf{4.57}/\textbf{75.0} \\
\cmidrule(l){2-6}
& \textit{Avg.} & 26.2/5.72/62.5 & 20.4/7.20/39.4 & \underline{17.1}/\underline{5.30}/\underline{63.1} & \textbf{13.2}/\textbf{4.56}/\textbf{80.0} \\
\midrule
\rowcolor{blue!10}
\multicolumn{6}{@{}l}{\textit{\textbf{Manipulation}} \quad SR (\%) $\uparrow$} \\
1 & Stack six-cup pyramid & \textbf{75.0} & 35.0 & 45.0 & \underline{72.5} \\
2 & Insert tubes into metal rack & \textbf{77.5} & 35.0 & 42.5 & \underline{67.5} \\
3 & Insert tubes into plastic rack & \textbf{65.0} & 17.5 & 35.0 & \underline{60.0} \\
4 & Sort medicine boxes and parts & \underline{62.5} & 47.5 & 42.5 & \textbf{65.0} \\
5 & Classify colored blocks & \underline{80.0} & 55.0 & 67.5 & \textbf{87.5} \\
6 & Place objects into tray & \underline{87.5} & 67.5 & 70.0 & \textbf{90.0} \\
7 & Prepare coffee, long-horizon & \textbf{52.5} & 0.0 & 7.5 & \underline{40.0} \\
\cmidrule(l){2-6}
& \textit{Avg.} & \textbf{71.4} & 36.8 & 44.3 & \underline{68.9} \\
\bottomrule
\end{tabularx}
\end{table}

\textbf{Zero-shot transfer from MAP-Data to real robots.}
\label{app:new-zeroshot}
This experiment tests whether MAP-Data alone teaches real-robot docking. Each model is trained only on MAP-Data for 70,000 steps on 32 NVIDIA H200 GPUs and is deployed directly on the Cobot Magic and the AgiBot G2. We refer to UniWAM trained in this way as UniWAM-Sim. The baselines use the same data and budget and otherwise follow their official recipes. Evaluation uses the 13 real MAP Navigation tasks and the protocol above, with 40 trials per task on each embodiment, and the errors are averaged over the two embodiments as for the real-data models. No real-robot data, room, or target instance is used in training, so every real-robot test is unseen. Table~\ref{tab:new-zeroshot} shows that UniWAM-Sim stops 18.7\,cm and $6.34^\circ$ from the MAP on average. Without any real-robot data, its position error is lower than that of $\pi_{0.5}$ trained on the real demonstrations, although its heading error remains higher than the $5.15^\circ$ of that model. The baselines transfer less well. The position errors of $\pi_{0.5}$ and GR00T N1.7 are 2.4 and 1.9 times those of their real-data versions in Table~\ref{tab:real-world-eval}, and FastWAM, which shares the video backbone of UniWAM, remains 12.1\,cm behind UniWAM-Sim. This shows that navigation learned from MAP-Data transfers to real robots, and the gap to FastWAM is consistent with the stream-specific design and image-plane supervision.

\begin{table}[t]
\caption{Zero-shot real-robot MAP Navigation of models trained only on MAP-Data, averaged over the Cobot Magic and the AgiBot G2. The bottom rows are models trained on real demonstrations, taken from Table~\ref{tab:real-world-eval}.}
\label{tab:new-zeroshot}
\centering
\small
\setlength{\tabcolsep}{8pt}
\begin{tabular}{@{}llcc@{}}
\toprule
Model & Training data & $E_{\mathrm{pos}}$ (cm) & $E_{\mathrm{head}}$ ($^\circ$) \\
\midrule
$\pi_{0.5}$ & MAP-Data only & 55.1 & 11.4 \\
GR00T N1.7 & MAP-Data only & 35.0 & 12.9 \\
FastWAM & MAP-Data only & 30.8 & 8.61 \\
UniWAM-Sim & MAP-Data only & 18.7 & 6.34 \\
\midrule
\multicolumn{4}{@{}l}{\textit{Reference: trained on real demonstrations}} \\
$\pi_{0.5}$ & real task data & 23.3 & 5.15 \\
UniWAM-Base & real task data & 15.1 & 4.91 \\
\bottomrule
\end{tabular}
\end{table}

\textbf{Statistical reliability.}
\label{app:new-stats}
Table~\ref{tab:new-ci} reports 95\% Wilson intervals for the real-robot success rates. Each task has 40 trials, so the category means of Table~\ref{tab:real-world-eval} equal the success rates pooled over the 160 Mobile Manipulation trials and the 280 Manipulation trials of Table~\ref{tab:real-world-task-eval}. On Mobile Manipulation, the intervals of UniWAM-Scaled and $\pi_{0.5}$ do not overlap. On Manipulation, the intervals of UniWAM-Scaled and $\pi_{0.5}$ overlap, as do those of UniWAM-Base and $\pi_{0.5}$ on Mobile Manipulation. Hence, this comparison shows no clear difference between these pairs, while UniWAM-Base still docks more accurately than $\pi_{0.5}$. The intervals of UniWAM-Base and its real-robot ablations also overlap, so the ablation conclusions rest on the error metrics together with the success rates. These intervals treat trials as independent and ignore the shared initial configurations, so they are conservative for paired comparisons.

\begin{table}[t]
\caption{Real-robot success rates with 95\% Wilson intervals, pooled over 160 Mobile Manipulation trials and 280 Manipulation trials per method, matching the category means of Table~\ref{tab:real-world-eval}. Brackets give the interval bounds.}
\label{tab:new-ci}
\centering
\small
\setlength{\tabcolsep}{8pt}
\begin{tabular}{@{}lcc@{}}
\toprule
Model & Mobile Manipulation SR (\%) & Manipulation SR (\%) \\
\midrule
$\pi_{0.5}$ & 62.5 {\scriptsize [54.8, 69.6]} & 71.4 {\scriptsize [65.9, 76.4]} \\
GR00T N1.7 & 39.4 {\scriptsize [32.1, 47.1]} & 36.8 {\scriptsize [31.4, 42.6]} \\
UniWAM-Base & 63.1 {\scriptsize [55.4, 70.2]} & 44.3 {\scriptsize [38.6, 50.1]} \\
UniWAM-Scaled & 80.0 {\scriptsize [73.1, 85.5]} & 68.9 {\scriptsize [63.3, 74.1]} \\
\midrule
w/o navigation auxiliary supervision & 56.9 {\scriptsize [49.1, 64.3]} & 43.6 {\scriptsize [37.9, 49.4]} \\
w/o manipulation auxiliary supervision & 58.8 {\scriptsize [51.0, 66.1]} & 39.3 {\scriptsize [33.7, 45.1]} \\
w/o independent sampling & 60.0 {\scriptsize [52.3, 67.3]} & 41.8 {\scriptsize [36.2, 47.6]} \\
\bottomrule
\end{tabular}
\end{table}

\subsection{Comparison of Unified Architectures}
\label{sec:architecture-results}

\begin{figure}[t]
\centering
\includegraphics[width=0.62\textwidth]{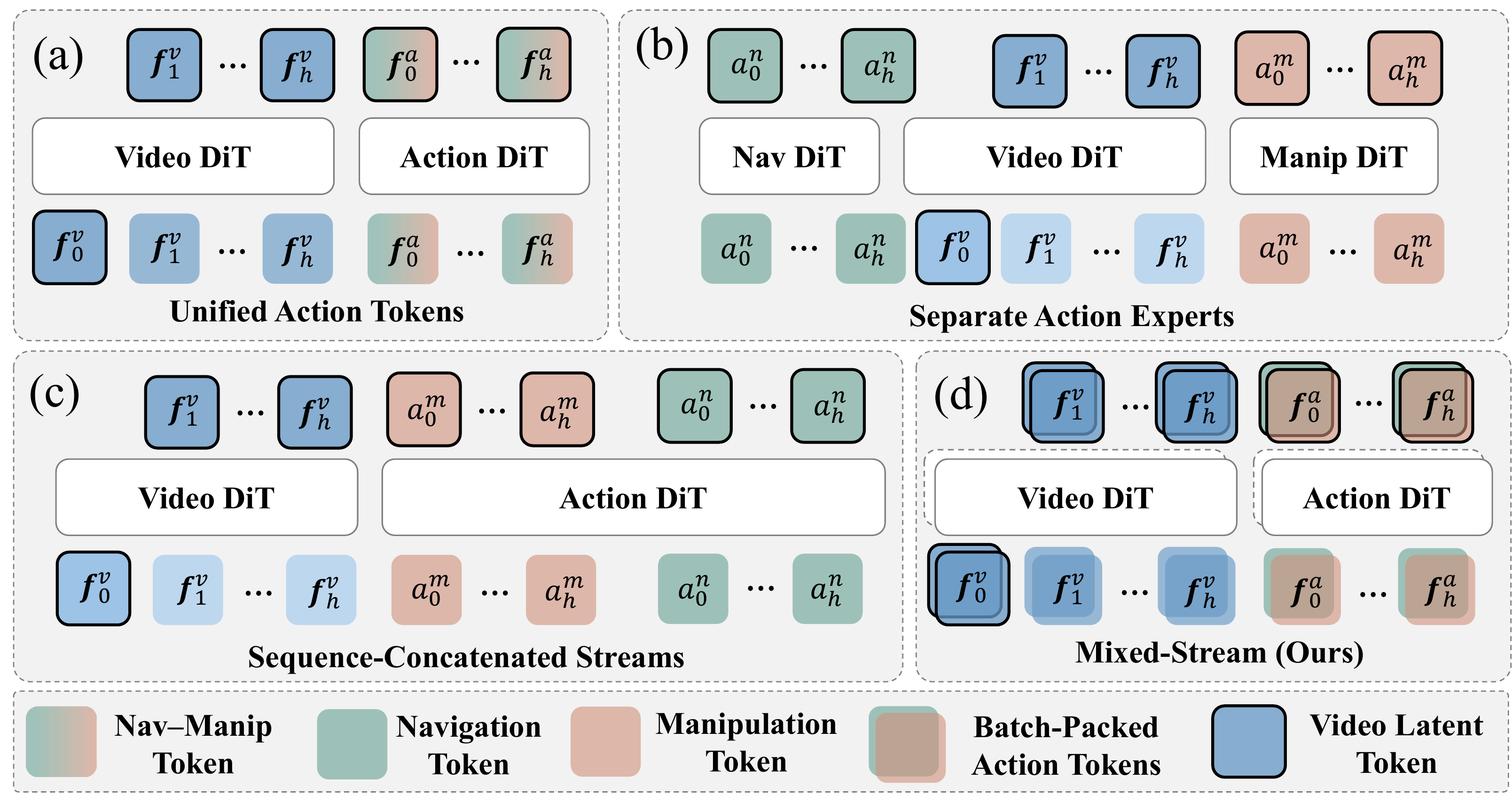}
\caption{Unified mobile-manipulation architectures compared in Table~\ref{tab:architecture-comparison}: (a) unified action tokens, (b) separate navigation and manipulation action DiTs, (c) streams concatenated along the sequence axis, and (d) Mixed-Stream (ours), with streams in separate batch rows of shared video and action DiTs and independent stream sampling.}
\label{fig:stream-ablation}
\end{figure}

Fig.~\ref{fig:stream-ablation} and Table~\ref{tab:architecture-comparison} compare four ways of unifying navigation and manipulation. All variants share the video DiT initialization, training data, optimization budget, evaluation protocol, and image-plane auxiliary targets, and they are evaluated on the four Mobile Manipulation tasks with the UniWAM-Base data. They differ only in how the two behaviors are represented, processed, and sampled, which isolates the effect of the stream design.
\label{app:architecture-comparison}

\textbf{(a) Unified Action Tokens.} Following $\pi_{0.5}$~\citep{black2025pi05}, GR00T N1.7~\citep{bjorck2025gr00t}, and FastWAM~\citep{yuan2026fastwam}, each action token contains both the navigation and manipulation dimensions, and one action DiT predicts the combined command.

\textbf{(b) Separate Action Experts.} Navigation and manipulation use separate action DiTs that share the video DiT, and each action DiT predicts the command of its own behavior. This variant represents behavior-specific action experts in general and is implemented within our framework, not as a reimplementation of a specific prior model. It therefore has more parameters than the mixed-stream design, which uses one action DiT.

\textbf{(c) Sequence-Concatenated Streams.} This is a controlled variant, not a reimplementation of a prior method. Navigation and manipulation action sequences are concatenated into one token sequence and processed by one action DiT, so the two behaviors occupy different positions of one row and attend to each other.

\textbf{(d) Mixed-Stream.} In the proposed design, the two behaviors keep separate visual inputs, action encoders, and output heads, share the video and action DiTs, and occupy separate batch rows.

The first three variants require navigation and manipulation targets in every sample, so they are trained with joint sampling. We evaluate the mixed-stream design with both joint and independent sampling to separate the effect of the architecture from that of sampling, since this is the only change between them.

\begin{table}[t]
\centering
\caption{Architecture and sampling comparison on the four Mobile Manipulation tasks. (a) to (c) use joint sampling. (d) uses joint or independent sampling, and the latter is UniWAM-Base as reported in Table~\ref{tab:real-world-eval}.}
\label{tab:architecture-comparison}
\small
\setlength{\tabcolsep}{8pt}
\begin{tabular}{@{}lccccc@{}}
\toprule
& \multicolumn{3}{c}{\textbf{Alternative architectures}} & \multicolumn{2}{c}{\textbf{(d) Mixed-stream}} \\
\cmidrule(lr){2-4}\cmidrule(l){5-6}
\textbf{Metric} & (a) Unified tokens & (b) Sep. experts & (c) Seq. concat. & Joint & \textbf{Indep.} \\
\midrule
$E_{\mathrm{pos}}$ (cm) $\downarrow$ & 22.0 & 19.7 & 22.1 & 18.6 & \textbf{17.1} \\
$E_{\mathrm{head}}$ ($^\circ$) $\downarrow$ & 7.49 & 6.06 & 6.78 & 5.87 & \textbf{5.30} \\
SR (\%) $\uparrow$ & 46.9 & 54.4 & 60.0 & 60.0 & \textbf{63.1} \\
\bottomrule
\end{tabular}
\end{table}

\textbf{Architecture.} Under joint sampling, the mixed-stream design obtains the lowest docking errors, 18.6\,cm and $5.87^\circ$, and ties Sequence-Concatenated Streams in success at 60.0\%. Compared with Unified Action Tokens, it reduces position and heading errors by 15.5\% and 21.6\% and improves success by 13.1 percentage points. Compared with Separate Action Experts, it lowers position and heading errors by 5.6\% and 3.1\% and raises success by 5.6 percentage points, despite a single shared action DiT. Relative to Sequence-Concatenated Streams, it reduces position and heading errors by 15.8\% and 13.4\% at the same success rate of 60.0\%, so separating the streams into batch rows helps docking.

\textbf{Sampling.} In the mixed-stream design, independent sampling further reduces the errors from 18.6\,cm and $5.87^\circ$ to 17.1\,cm and $5.30^\circ$ and increases success from 60.0\% to 63.1\%. Both design choices reduce docking errors, and together they give the highest success of all five variants in the table.

\subsection{Ablation Studies}
\label{sec:uniwam-ablations}
\label{app:additional-ablations}

Each ablation changes one factor of UniWAM-Base with the data, training budget, and evaluation protocol fixed. Table~\ref{tab:mapbench-ablation} reports the MAP-Bench ablations, and Table~\ref{tab:real-ablation} reports the real-robot ablations.

\begin{table}[t]
\caption{Ablations of UniWAM-Base on MAP-Bench. The first row is the full model.}
\label{tab:mapbench-ablation}
\centering
\small
\setlength{\tabcolsep}{5pt}
\begin{tabular}{@{}lrrr r||rrr r@{}}
\toprule
& \multicolumn{3}{c}{\textbf{$E_{\mathrm{pos}}$ (cm) $\downarrow$}} & \multicolumn{1}{c||}{\multirow{2}{*}{\textbf{Overall}}} & \multicolumn{3}{c}{\textbf{$E_{\mathrm{head}}$ ($^\circ$) $\downarrow$}} & \multirow{2}{*}{\textbf{Overall}} \\
\cmidrule(lr){2-4}\cmidrule(lr){6-8}
\textbf{Variant} & \textbf{Short} & \textbf{Mid} & \textbf{Long} & & \textbf{Short} & \textbf{Mid} & \textbf{Long} & \\
\midrule
UniWAM-Base & 7.71 & 8.85 & 21.0 & 12.5 & 2.37 & 3.06 & 4.29 & 3.24 \\
\midrule
Velocity actions $(v,\omega)$ & 8.30 & 9.57 & 30.9 & 16.3 & 3.03 & 3.29 & 6.19 & 4.17 \\
w/o causal mask & 8.83 & 10.1 & 29.0 & 16.0 & 3.21 & 3.50 & 5.49 & 4.07 \\
w/o target box & 7.74 & 9.29 & 29.5 & 15.5 & 2.43 & 3.02 & 5.14 & 3.53 \\
w/o image-plane MAP & 8.05 & 8.40 & 22.5 & 13.0 & 2.63 & 2.82 & 4.72 & 3.39 \\
w/o both auxiliary targets & 9.31 & 11.6 & 29.1 & 16.7 & 4.45 & 4.91 & 6.76 & 5.37 \\
\bottomrule
\end{tabular}
\end{table}

\textbf{Pose targets.} Replacing the relative pose target $(\Delta x,\Delta y,\Delta\theta)$ with velocity commands $(v,\omega)$ increases the Overall errors from 12.5\,cm and $3.24^\circ$ to 16.3\,cm and $4.17^\circ$. The position error rises only from 7.71 to 8.30\,cm at short range but from 21.0 to 30.9\,cm at long range. The degradation grows with distance, consistent with velocity errors that accumulate over time and with the bimodal velocity labels discussed in Sec.~\ref{app:action-representations}.

\textbf{Causal mask.} Removing the causal mask increases the Overall errors to 16.0\,cm and $4.07^\circ$ and the long-range position error from 21.0 to 29.0\,cm. The mask lets later actions condition on the clean committed prefix without leaking information from later noisy actions backward, which matters most over long approaches with many successive chunks.

\textbf{Navigation auxiliary targets.} The two navigation targets have different distance-dependent effects. Removing the target box barely changes the short-range errors but raises the long-range errors from 21.0\,cm and $4.29^\circ$ to 29.5\,cm and $5.14^\circ$, suggesting that the box helps localize distant, small targets. Removing the projected MAP, which marks where the base should stop, instead raises the short-range errors more than removing the box, from 7.71\,cm and $2.37^\circ$ to 8.05\,cm and $2.63^\circ$. Removing both raises the Overall errors to 16.7\,cm and $5.37^\circ$, above those of GR00T N1.7 and $\pi_{0.5}$, which suggests that the two targets have complementary effects on docking.

\begin{table}[t]
\caption{Ablations of UniWAM-Base on real robots. The first row is the full model.}
\label{tab:real-ablation}
\centering
\small
\setlength{\tabcolsep}{5pt}
\begin{tabular}{@{}lcc||ccc||c@{}}
\toprule
& \multicolumn{2}{c||}{\textbf{MAP Navigation}} & \multicolumn{3}{c||}{\textbf{Mobile Manipulation}} & \textbf{Manipulation} \\
\cmidrule(lr){2-3}\cmidrule(lr){4-6}\cmidrule(lr){7-7}
\textbf{Variant} & \textbf{$E_{\mathrm{pos}}$ (cm) $\downarrow$} & \textbf{$E_{\mathrm{head}}$ ($^\circ$) $\downarrow$} & \textbf{$E_{\mathrm{pos}}$ (cm) $\downarrow$} & \textbf{$E_{\mathrm{head}}$ ($^\circ$) $\downarrow$} & \textbf{SR (\%) $\uparrow$} & \textbf{SR (\%) $\uparrow$} \\
\midrule
UniWAM-Base & 15.1 & 4.91 & 17.1 & 5.30 & 63.1 & 44.3 \\
\midrule
w/o navigation auxiliary supervision & 17.6 & 6.84 & 19.9 & 7.21 & 56.9 & 43.6 \\
w/o manipulation auxiliary supervision & 15.4 & 5.18 & 17.8 & 5.61 & 58.8 & 39.3 \\
w/o independent sampling & 16.8 & 5.62 & 18.6 & 5.87 & 60.0 & 41.8 \\
\bottomrule
\end{tabular}
\end{table}

\textbf{Auxiliary supervision on real robots.} Removing navigation auxiliary supervision changes MAP Navigation from 15.1\,cm and $4.91^\circ$ to 17.6\,cm and $6.84^\circ$ and lowers Mobile Manipulation success from 63.1\% to 56.9\%. Removing manipulation auxiliary supervision leaves navigation close to the full model but lowers Manipulation success from 44.3\% to 39.3\% and Mobile Manipulation success from 63.1\% to 58.8\%. Each auxiliary target has its largest effect on its own stream, and Figs.~\ref{fig:mapbench-aux} and~\ref{fig:eef-aux} visualize these predictions.

\textbf{Independent sampling.} Removing independent sampling worsens all six real-robot metrics, raising the MAP Navigation errors to 16.8\,cm and $5.62^\circ$ and lowering Mobile Manipulation success to 60.0\%. Independent sampling thus improves both the navigation and the manipulation stream without any extra data.

\textbf{Learnable tokens.}
\label{app:learned-stream-tokens}
Table~\ref{tab:learned-stream-token-ablation} varies the number of learnable tokens per token bank over $\{0,4,8\}$, using the same count for all four banks. Removing the tokens degrades all six metrics. The MAP Navigation errors increase from 15.1\,cm and $4.91^\circ$ to 16.2\,cm and $5.38^\circ$, and Mobile Manipulation success drops from 63.1\% to 61.9\%. Explicit token identity thus helps the shared DiTs distinguish the two streams. Increasing the count to eight changes each success rate by at most one trial, from 63.1\% to 63.8\% and from 44.3\% to 44.6\%, and gives mixed error changes, so we use four tokens per bank.

\begin{table}[t]
\caption{Real-robot results for the number of learnable tokens per bank. $\dagger$ marks the default.}
\label{tab:learned-stream-token-ablation}
\centering
\small
\setlength{\tabcolsep}{5pt}
\begin{tabular}{@{}lcc||ccc||c@{}}
\toprule
& \multicolumn{2}{c||}{\textbf{MAP Navigation}} & \multicolumn{3}{c||}{\textbf{Mobile Manipulation}} & \textbf{Manipulation} \\
\cmidrule(lr){2-3}\cmidrule(lr){4-6}\cmidrule(lr){7-7}
\textbf{Tokens per bank} & \textbf{$E_{\mathrm{pos}}$ (cm) $\downarrow$} & \textbf{$E_{\mathrm{head}}$ ($^\circ$) $\downarrow$} & \textbf{$E_{\mathrm{pos}}$ (cm) $\downarrow$} & \textbf{$E_{\mathrm{head}}$ ($^\circ$) $\downarrow$} & \textbf{SR (\%) $\uparrow$} & \textbf{SR (\%) $\uparrow$} \\
\midrule
0 & 16.2 & 5.38 & 17.8 & 5.48 & 61.9 & 43.6 \\
4, ours$^\dagger$ & 15.1 & 4.91 & 17.1 & 5.30 & 63.1 & 44.3 \\
8 & 14.8 & 5.02 & 17.3 & 5.24 & 63.8 & 44.6 \\
\bottomrule
\end{tabular}
\end{table}

\textbf{Manipulation coordinate frame.}
\label{app:coordinate-frame}
Table~\ref{tab:uniwam-frames} compares camera-frame and robot-base end-effector targets under multi-embodiment co-training. Camera-frame targets improve success on each reported embodiment and task group and raise the unweighted mean from 48.5\% to 55.2\%. Following the alignment principle of Qwen-RobotManip~\citep{yuan2026qwen-robotmanip}, expressing end-effector motion in the current camera frame makes visually similar motions numerically similar across embodiments. The same visual change in the egocentric image then corresponds to similar action targets, regardless of the base frame or arm geometry of each robot, which lets data from different embodiments reinforce one another. This comparison covers co-training on these embodiments and does not evaluate transfer to unseen robots.

\begin{table}[t]
\caption{Manipulation target frame in UniWAM-Base under multi-embodiment co-training. Values are success rates in percent. Cobot Magic uses the Mobile Manipulation tasks, the dual Franka uses Manipulation tasks 1 to 3 and 7, and the AgileX single arm uses Manipulation task 6. Mean is the unweighted mean over the three reported embodiment and task groups.}
\label{tab:uniwam-frames}
\centering
\small
\setlength{\tabcolsep}{8pt}
\begin{tabular}{@{}lcccc@{}}
\toprule
\textbf{Target frame} & \textbf{Cobot Magic} & \textbf{Dual Franka} & \textbf{AgileX single arm} & \textbf{Mean} \\
\midrule
Robot base & 55.0 & 28.1 & 62.5 & 48.5 \\
\textbf{Camera, ours} & \textbf{63.1} & \textbf{32.5} & \textbf{70.0} & \textbf{55.2} \\
\bottomrule
\end{tabular}
\end{table}

\subsection{Qualitative Results}
\label{sec:qualitative-results}

\begin{figure}[p]
\centering
\includegraphics[width=\textwidth,height=0.9\textheight,keepaspectratio]{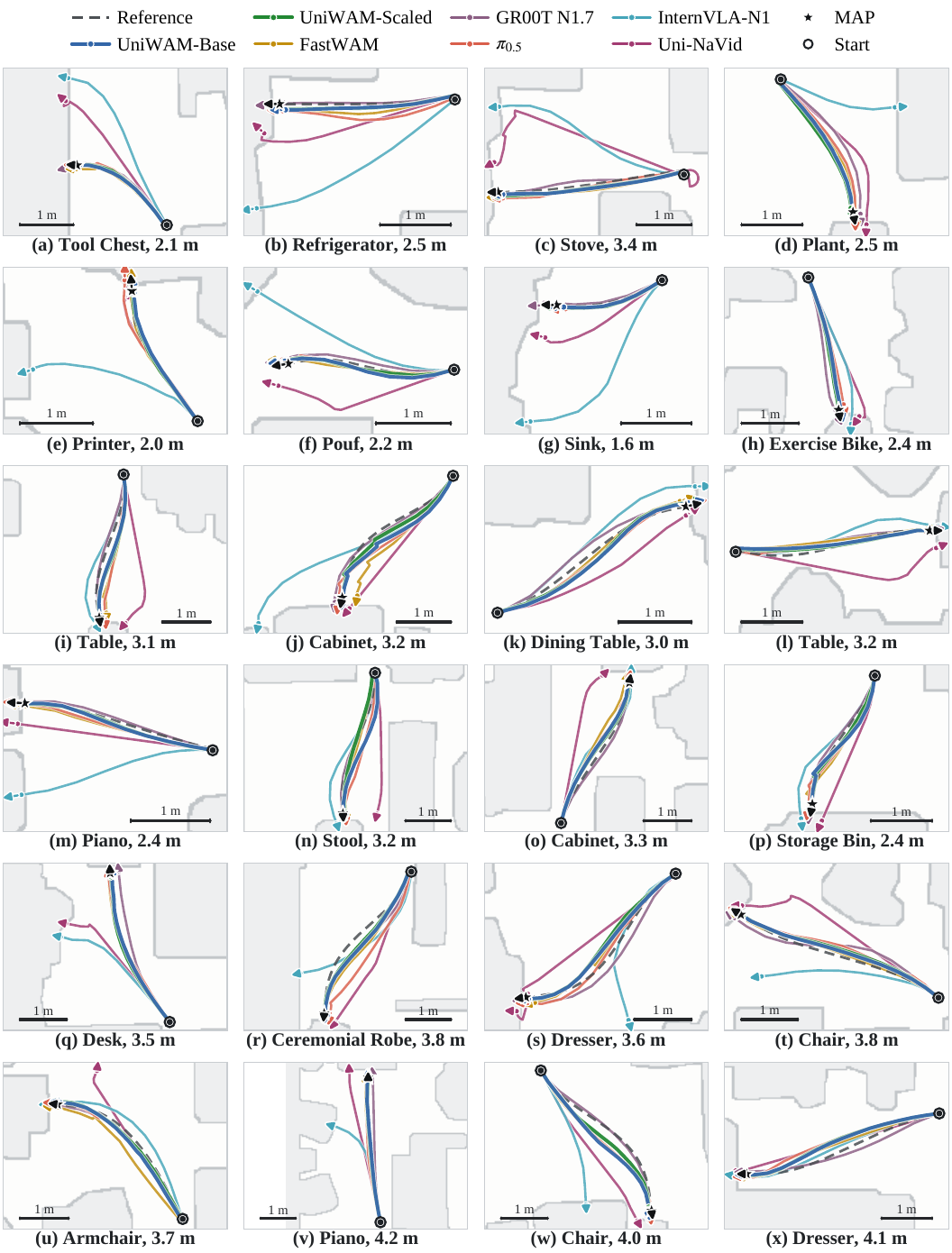}
\caption{MAP-Bench trajectories of all evaluated methods on 24 randomly selected scenes. Subcaptions give the target category and initial distance. Stars mark the reference MAP, circles the start pose, and arrowheads the terminal heading. In the last two rows, every episode starts at least 3.5\,m from the MAP, which falls in the long range.}
\label{fig:mapbench-qualitative-24}
\end{figure}

\textbf{Trajectories.} Fig.~\ref{fig:mapbench-qualitative-24} compares the trajectories of all methods on 24 randomly selected MAP-Bench scenes with initial distances from 1.6 to 4.2\,m. The two VLN baselines show the typical failures of semantic navigation. InternVLA-N1 heads to a wrong region in many displayed scenes, for example in panels a to g. Uni-NaVid often reaches the vicinity of the target but stops beside it, as in panels b, f, l, and n. The VLA and WAM baselines reach the target region in most displayed scenes but show visible path deviations or terminal offsets, such as FastWAM in panels j and u and $\pi_{0.5}$ in panel r. Both UniWAM variants follow the reference path closely and end at the MAP in most displayed scenes. The contrast is largest in the last two rows, where all initial distances are at least 3.5\,m. There, the VLN baselines fail in most displayed scenes and the other baselines deviate more often, whereas UniWAM remains close to the reference path, in agreement with the long-range gains of both UniWAM variants in Table~\ref{tab:mapbench-baselines}.

\begin{figure}[t]
\centering
\includegraphics[width=\textwidth]{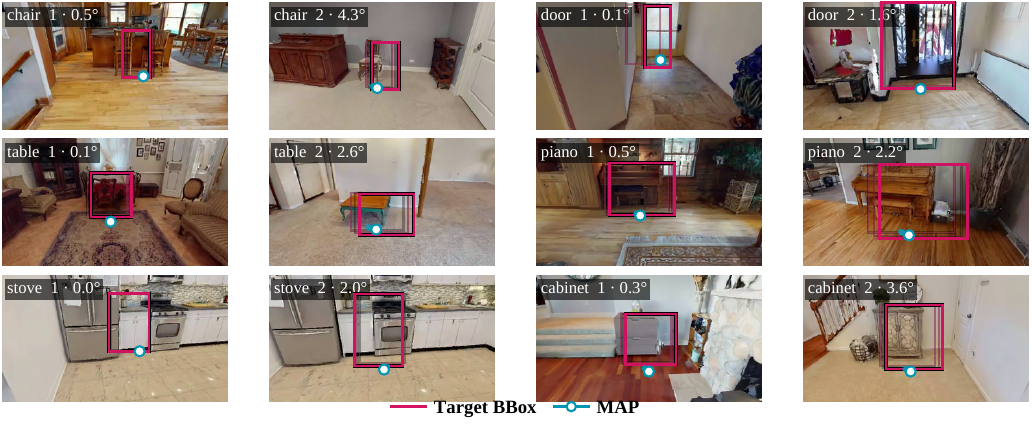}
\caption{Navigation auxiliary predictions of UniWAM on MAP-Bench, overlaid over consecutive frames of each episode. Magenta boxes are the predicted target boxes, and cyan points are the predicted MAP positions projected into the image, both predicted jointly with the actions by the navigation stream of UniWAM at every step of the approach.}
\label{fig:mapbench-aux}
\end{figure}

\textbf{Auxiliary predictions.} Fig.~\ref{fig:mapbench-aux} overlays the target boxes and projected MAP positions that UniWAM predicts over consecutive frames of MAP-Bench episodes. The boxes tightly enclose the target across object categories, scenes, and viewing distances, from large targets that fill much of the view, such as the stove and the door, to smaller targets such as the chairs. The boxes from consecutive frames nearly coincide, so the predictions remain stable over time rather than jittering from frame to frame. The projected MAP lies on the floor directly in front of the manipulation-facing side of the target, at the point where the base should stop. These observations match the ablation in Sec.~\ref{sec:uniwam-ablations}. The box keeps the target localized throughout the approach, and the MAP point marks the terminal pose.

\begin{figure}[t]
\centering
\includegraphics[width=0.85\textwidth]{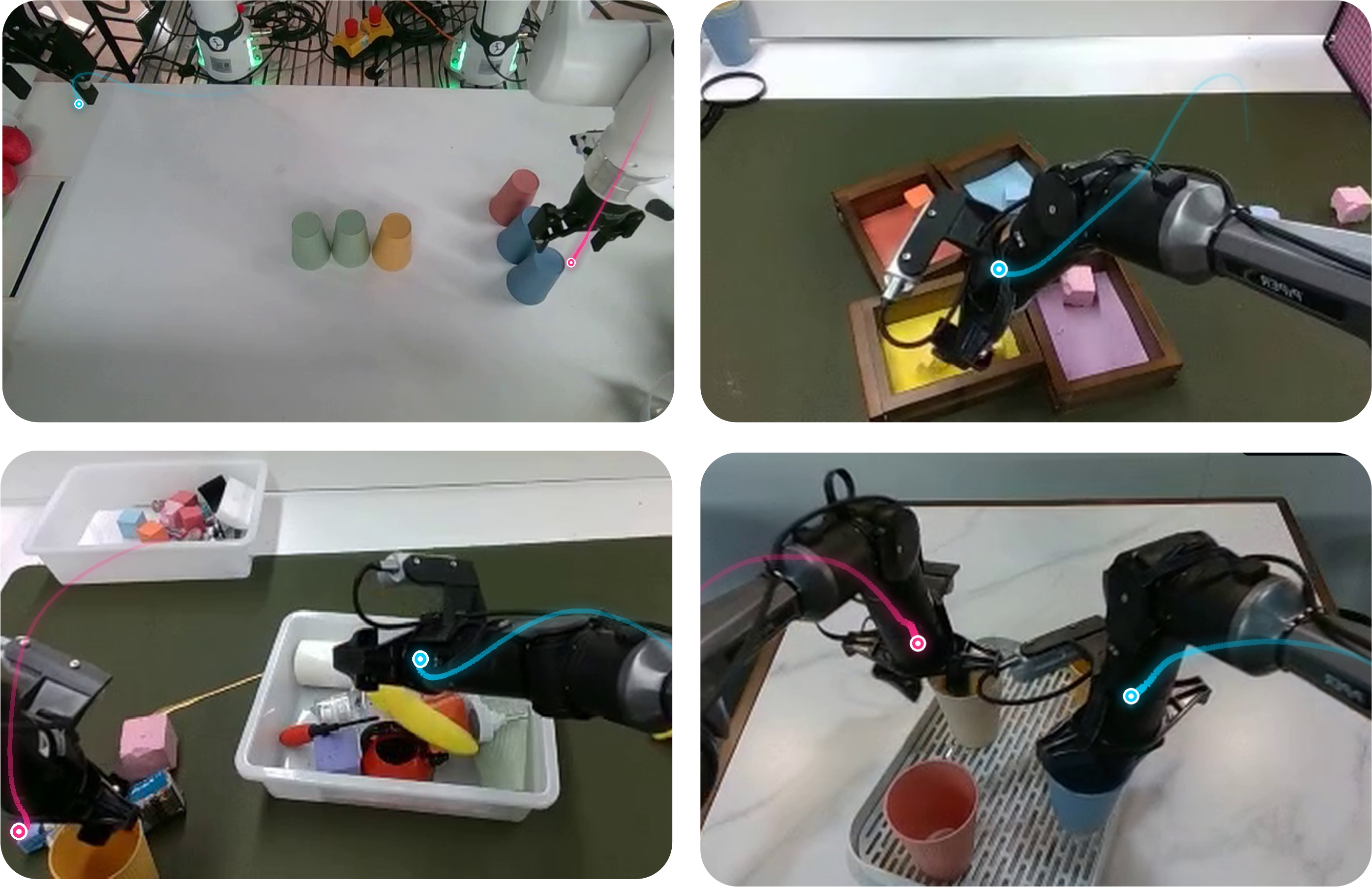}
\caption{Alignment between predicted 2D end-effector tracks and executed 3D actions on several tasks and two embodiments. Curves show the tracks of the two arms predicted in the previous action chunk, and circles mark the positions reached after that chunk.}
\label{fig:eef-aux}
\end{figure}

\textbf{End-effector predictions.} Fig.~\ref{fig:eef-aux} examines whether the predicted 2D end-effector tracks agree with the executed 3D actions. Each curve is the track predicted for one arm in the previous action chunk, and each circle marks the actual end-effector position after the robot executes that chunk. Across both arms, several tasks, and two embodiments, the circles lie on the predicted tracks at the grippers. In other words, the image-plane tracks agree with where the camera-frame actions actually move the end effectors. The manipulation stream has therefore learned a consistent mapping from its 3D end-effector actions to their pixel locations in the main camera view of each embodiment.

\subsection{Failure Cases}
\label{app:failure-cases}

\begin{figure}[H]
\centering
\includegraphics[width=0.27\linewidth]{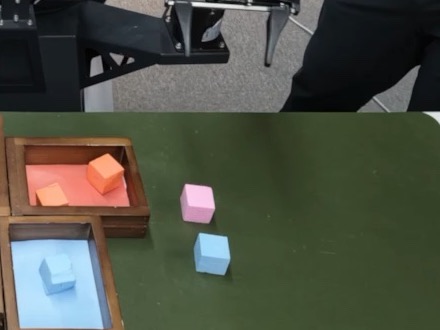}\hfill
\includegraphics[width=0.27\linewidth]{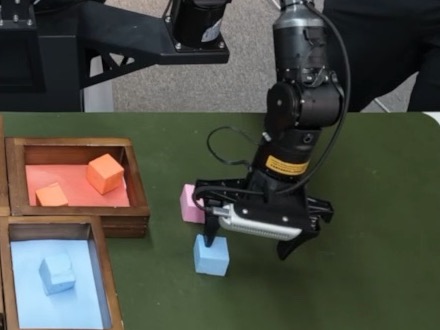}\hfill
\includegraphics[width=0.27\linewidth]{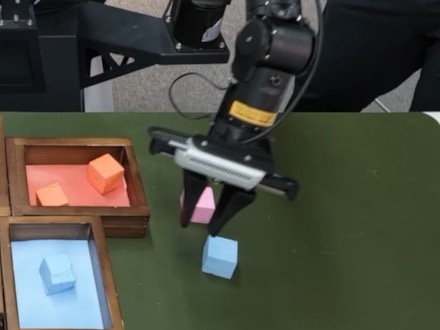}
\caption{Failure case in the color-sorting task. From left to right, a pink and a blue cube stand next to each other, and the gripper descends toward the point between them and closes on the empty gap instead of a cube.}
\label{fig:new-failure-case}
\end{figure}

\textbf{Grasping between adjacent objects.} Fig.~\ref{fig:new-failure-case} shows a failure of UniWAM in the color-sorting task. A pink and a blue cube stand next to each other on the table, and either may be picked next. Instead of committing to one of them, the gripper descends toward the point between the two cubes and closes on the gap. The demonstrations contain grasps of either cube, so the action distribution has two nearby modes. When the modes are close, the deterministic ten-step sampler can produce an action between them rather than on one of them. We observed this failure when candidate objects were close together, and specifying the target object in the instruction may reduce the ambiguity.

\textbf{Residual base motion.} During the manipulation stage, the navigation stream predicts motions that are close to but not exactly zero, which can move the base slightly. The near-zero rule suppresses it for all methods.

\FloatBarrier
\section{Conclusion}
\label{sec:conclusion}

We presented UniWAM, a mixed-stream world-action model that treats navigation and manipulation as two independently sampled control streams over shared video and action DiTs, and the Manipulation Anchor Pose (MAP), an explicit target for manipulation-ready navigation supported by over 1.5 million automatically generated MAP-Data episodes. Independent sampling lets UniWAM learn directly from navigation-only and manipulation-only data, and image-plane auxiliary supervision grounds the actions of both streams in the egocentric image. On MAP-Bench, UniWAM trained with MAP-Data reduces position and heading errors by 30.1\% and 44.0\% relative to the strongest baselines. On 24 real-robot tasks across four embodiments, a single model trained with additional data achieves the most accurate docking and the highest mobile-manipulation success while remaining competitive in manipulation. 

\textbf{Future work.} We will conduct large-scale real-robot pretraining of the UniWAM architecture, combining diverse real-robot demonstrations with MAP-Data, and study how the mixed-stream design scales with robot data.

\FloatBarrier
\bibliographystyle{plainnat}
\bibliography{main}

\end{document}